\documentclass[lettersize,journal]{IEEEtran}
\usepackage{amsmath,amsfonts}
\usepackage{algorithmic}
\usepackage{algorithm}
\usepackage{array}
\usepackage[caption=false,font=normalsize,labelfont=sf,textfont=sf]{subfig}
\usepackage{textcomp}
\usepackage{stfloats}
\usepackage{url}
\usepackage{verbatim}
\usepackage{graphicx}
\usepackage{cite}
\usepackage{color}
\usepackage{makecell}
\usepackage{multirow}
\usepackage{booktabs}
\usepackage{hyperref}
\usepackage{threeparttable}
\usepackage{enumerate}
\begin{document}

\title{Consecutive Pretraining: A Knowledge Transfer Learning Strategy with Relevant Unlabeled Data for Remote Sensing Domain}

\author{Tong Zhang,~\IEEEmembership{Student Member,~IEEE,} Peng Gao,~\IEEEmembership{Member,~IEEE,} Hao Dong,~\IEEEmembership{Member,~IEEE,} \\Yin Zhuang,~\IEEEmembership{Member,~IEEE,} Guanqun Wang,~\IEEEmembership{Student Member,~IEEE,} Wei Zhang,~\IEEEmembership{Student Member,~IEEE,} and \\He Chen,~\IEEEmembership{Member,~IEEE}
\thanks{This work was supported by the Chang Jiang Scholars Program under grant T2012122, by the Civil Aviation Program under grant B0201, in part by the Space based on orbit real-time processing technology under grant 2018-JCJQ-ZQ-046, in part by the National Science Foundation for Young Scientists of China under Grant 62101046 as well as by the National Natural Science Foundation of China (No. 62136001)\\(Corresponding Author: Y. Zhuang: zhuangyin640829@163.com)\\T. Zhang, Y. Zhuang, G. Wang and H. Chen are with Beijing Key Laboratory of Embedded Real-time Information Processing Technology, Beijing Institute of Technology, Beijing 100081, China;\\P. Gao and W. Zhang are with the Shang Hai AI laboratory, Shanghai 100024, China;\\H. Dong is with the Center on Frontiers of Computing Studies (CFCS), Peking University, Beijing 100871, China.}}

\markboth{Journal of \LaTeX\ Class Files,~Vol.~14, No.~8, August~2021}%
{Shell \MakeLowercase{\textit{et al.}}: A Sample Article Using IEEEtran.cls for IEEE Journals}

\maketitle
\begin{abstract}
Currently, under supervised learning, a model pretrained by a large-scale nature scene dataset and then fine-tuned on a few specific task labeling data is the paradigm that has dominated the knowledge transfer learning. It has reached the status of consensus solution for task-aware model training in remote sensing domain (RSD). Unfortunately, due to different categories of imaging data and stiff challenges of data annotation, there is not a large enough and uniform remote sensing dataset to support large-scale pretraining in RSD. Moreover, pretraining models on large-scale nature scene datasets by supervised learning and then directly fine-tuning on diverse downstream tasks seems to be a crude method, which is easily affected by inevitable labeling noise, severe domain gaps and task-aware discrepancies. Thus, in this paper, considering the self-supervised pretraining and powerful vision transformer (ViT) architecture, a concise and effective knowledge transfer learning strategy called ConSecutive PreTraining (CSPT) is proposed based on the idea of not stopping pretraining in natural language processing (NLP), which can gradually bridge the domain gap and transfer knowledge from the nature scene domain to the RSD. The proposed CSPT also can release the huge potential of unlabeled data for task-aware model training. Finally, extensive experiments are carried out on twelve datasets in RSD involving three types of downstream tasks (e.g., scene classification, object detection and land cover classification) and two types of imaging data (e.g., optical and SAR). The results show that by utilizing the proposed CSPT for task-aware model training, almost all downstream tasks in RSD can outperform the previous method of supervised pretraining-then-fine-tuning and even surpass the state-of-the-art (SOTA) performance without any expensive labeling consumption and careful model design.
\end{abstract}

\begin{IEEEkeywords}
 Knowledge Transfer Learning; Remote Sensing Domain; Self-Supervised Learning; Vision Transformer.
\end{IEEEkeywords}
\section{Introduction}
    \IEEEPARstart{W}{ith} the rapid development of remote sensing technology, there is a gradual accumulation of available earth observation imaging data, which can be used for 
    \begin{figure}
    \vspace{0.2cm}
	\centering 
	\setlength{\abovecaptionskip}{0.cm}
	\includegraphics[width=3.5in]{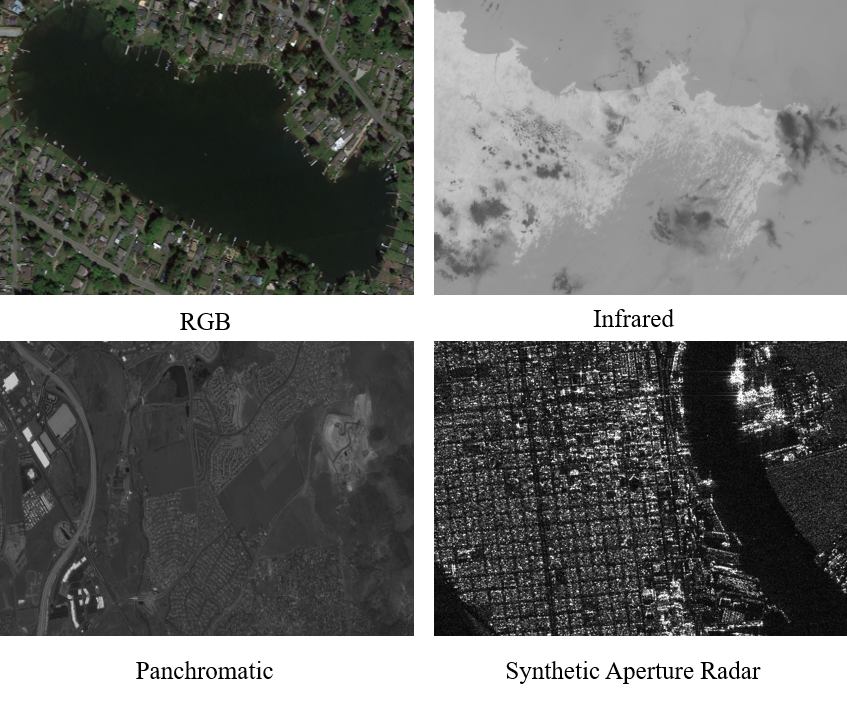}
	\caption{The different types of imaging data in remote sensing domain.}
	\vspace{-0.6cm}
    \end{figure}
    urban planning, resource investigation, military surveillance and rapid search \& rescue within a large-scale regions\cite{2019Seismic,2016Urban,Ozdarici2015Mapping,sadgrove2018real,reilly2010detection}. Consequently, how to convert a huge amount of unlabeled data into valid information to support practical applications, becomes a very important issue. Subsequently, various data-hungry models, such as convolutional neural networks (CNNs) \cite{jia2014caffe,simonyan2014very,szegedy2015going,krizhevsky2012imagenet,he2016deep,xie2017aggregated,zhang2020resnest,tan2019efficientnet} and vision transformers (ViTs) \cite{dosovitskiy2020image,liu2021swin,zhang2022vitaev2,gao2021container} have emerged with the evolution of deep learning techniques. These models promote the performance of intelligent interpretation and provide an opportunity to convert a large amount of unlabeled data into valid information by careful imaging data labeling and supervised learning. However, due to multi-payload, multi-platform and multi-revisit cycles of earth observation characteristics, a large amount of unlabeled imaging data (e.g., optical, infrared, synthetic aperture radar (SAR) and panchromatic) would have significant differences in imaging characteristics, spatial resolutions and imaging conditions as shown in Fig. 1. Thus, manual labeling of remote sensing data requires more professional knowledge, and it often suffers some labeling challenges in downstream tasks (e.g., object detection, scene classification and land cover classification), such as densely arranged tiny objects, ambiguous class definitions of some remote sensing scenes and complicated spatial semantic relations involving the signing lots of pixels. 
\begin{table*}
\vspace{0.2cm}
\centering
\caption{Natural scene and Remote sensing Datasets description.}
\renewcommand\arraystretch{1.3}
\setlength{\tabcolsep}{7pt}{
\begin{tabular}{ccccccccccc}
\toprule[1.2pt]
Domain                           & \multicolumn{2}{c}{Task}                                                                                     & Dataset       & Resolution(m) & Classes & \# Train  & \# Val & \# Test & Source & Year \\ \toprule[1pt]
\multirow{4}{*}{\begin{tabular}[c]{@{}c@{}}Natural \\ Scene\end{tabular}}         & \multicolumn{2}{c}{\multirow{2}{*}{Classification}}                                                          & ImageNet\cite{ILSVRC15}      & -             & 1,000    & 1,281,167 & 50,000 & 100,000 & Multi & 2009\\
                                 & \multicolumn{2}{c}{}                                                                                         & Place365\cite{zhou2017places}      & -             & 365     & 1,803,460 & -      & 36,500 & Multi & 2017\\ \cline{2-11}
                                 & \multicolumn{2}{c}{\multirow{2}{*}{Detection/Segmentation}} & COCO\cite{2014Microsoft}          & -             & 80      & 118,287   & 5,000  & 40,670 & Multi & 2014 \\
                                 & \multicolumn{2}{c}{}                                                                                         & PASCAL   VOC\cite{everingham2015pascal}  & -             & 20      & 11,530    & -      & -  & Multi  &2007   \\ \hline
\multirow{12}{*}{\begin{tabular}[c]{@{}c@{}}Remote \\ Sensing\end{tabular}} & \multirow{9}{*}{Optical}                          & \multirow{2}{*}{Classification}                          & AID\cite{2017AID}           & 0.5 to 8      & 30      & 2,000     & -      & 8,000  & Multi & 2017\\
                                 &                                                   &                                                          & NWPU-RESISC45\cite{2017Remote} & 0.2   to 30   & 45      & 9,450     & -      & 22,000 & Multi & 2016 \\ \cline{3-11}
                                 &                                                   & \multirow{3}{*}{Segmentation}                            & Potsdam\cite{isprs}       & 0.05          & 6       & 3,456     & -      & 2,016 &Single  & 2012\\
                                 &                                                   &                                                          & Vaihingen\cite{isprs}     & 0.09          & 6       & 344       & -      & 398   & Single  & 2012 \\
                                 &                                                   &                                                          & GID\cite{tong2020land}           & 0.8 to 3.24     & 15      & 4,368     & -      & 2,912   & Single & 2020 \\ \cline{3-11}
                                 &                                                   & \multirow{4}{*}{Detection}                               & DIOR\cite{2019Object}          & 0.5   to 30   & 20      & 5,862     & 5,863  & 11,738  & Multi & 2019 \\
                                 &                                                   &                                                          & NWPUVHR-10\cite{cheng2014multi}    & 0.5 to 2      & 10      & 1,479     & -      & 1,279  & Multi & 2014 \\
                                 &                                                   &                                                          & UCAS-AOD\cite{2015Orientation}      & -     & 2       & 6,489     & -      & 2,824    & Single & 2015 \\
                                 &                                                   &                                                          & HRSC2016\cite{liu2017high}      & 0.4 to 2      & 1       & 617       & -      & 438    & Single & 2017 \\ \cline{2-11}
                                 & \multirow{3}{*}{SAR}                              & Classification                                           & MSTAR\cite{1998Moving}         & 0.3           & 8       & 1,890     & -       & 7,576  & Single & 1998 \\ \cline{3-11}
                                 &                                                   & \multirow{2}{*}{Detection}                               
                                                                                        & SSDD\cite{li2017ship}          & 1   to 15     & 1       &  812    & -       & 348 & Multi & 2017 \\
                                &                                                   &    & HRSID\cite{wei2020hrsid}         & 0.5 to 3        & 1       &  3,642      & -       & 1,962   & Single  &2020  \\ \toprule[1.2pt]
\end{tabular}
}
\vspace{-0.6cm}
\end{table*}
    Therefore, as illustrated in Table I, in the remote sensing domain (RSD), there is not a large enough and uniform dataset to support large-scale pretraining. This restricts the fine-tuning performance on diverse downstream tasks with different types of imaging data. Naturally, the knowledge transfer learning from a large-scale nature scene dataset to a smaller scale remote sensing dataset is easily considered, and it can prevent the overfitting phenomenon when training a model from scratch with insufficient labeling data. Hence knowledge transfer learning is widely used for task-aware model training in RSD. 
    \par Facing to the data scarcity problem for data-hungry model pretraining and fine-tuning in RSD, many efforts have been made \cite{2017AID,2017Remote,isprs,tong2020land,2019Object,cheng2014multi,2015Orientation,liu2017high,1998Moving,wei2020hrsid,li2017ship,long2022aerial,ranjan2020build,liu2018semantic,chen2018end,wang2022empirical}. Related to remote sensing scene classification, G. Xia \textit{et al.}\cite{2017AID} constructed a benchmark called the aerial image dataset (AID) for remote sensing scene classification. In detail, they collected 10,000 RGB optical aerial images from the Google Earth platform involving 30 types of aerial scenes. Then, the knowledge transfer learning method was employed to evaluate the performance of CNN based architectures (e.g., CaffeNet\cite{jia2014caffe}, VGG-VD-16\cite{simonyan2014very} and GoogLeNet\cite{szegedy2015going}) on AID\cite{2017AID}. Under supervised learning, all CNNs are pretrained on ImageNet\cite{ILSVRC15} and then fine-tuned on AID\cite{2017AID}, which can obtain the best performance and show its great generalization ability compared with other traditional handcraft feature based methods. In addition, \cite{2017AID} also indicated that deeper CNNs pretrained on a large-scale dataset of nature scenes would learn more specific features oriented to natural image processing. Thus, they have worse performance than shallow layer CNNs for classifying aerial scenes. Next, G. Cheng \textit{et al.}\cite{2017Remote} built another benchmark for remote sensing scene classification called NWPU-RESISC45 (NR45), which contains 31,500 images covering 45 scene classes with 700 images in each class. In \cite{2017Remote}, the knowledge transfer learning method was also utilized for CNN-based architectures (e.g., VGGNet-16\cite{simonyan2014very}, GoogleNet\cite{szegedy2015going} and AlexNet\cite{krizhevsky2012imagenet}), and they were pretrained on a large-scale dataset of ImageNet\cite{ILSVRC15} and then fine-tuned on NR45\cite{2017Remote}. Then, the experimental results of \cite{2017Remote} proved that CNNs can provide superior scene classification performance on NR45\cite{2017Remote}, and the fine-tuning is a necessary step to further improve the performance for knowledge transfer learning. Then, Y. Long \textit{et al.}\cite{long2022aerial} set up a new larger scale aerial scene classification benchmark than \cite{2017AID} and \cite{2017Remote} called Million-AID (M-AID). The total number of image titles in M-AID\cite{long2022aerial} is 1,000,848 which involves 51 fine-grained labeling scene classes. The experiments of \cite{long2022aerial} demonstrated that CNN-based classification models trained from scratch would generate poor performance. However, when ImageNet\cite{ILSVRC15} or M-AID\cite{long2022aerial} is employed for the model pretraining step of knowledge transfer learning and then fine-tuning on AID\cite{2017AID} and NR45\cite{2017Remote}, one can attain competitive results. In addition, the study of \cite{long2022aerial} also proved that knowledge transfer learning from M-AID\cite{long2022aerial} to AID\cite{2017AID} or NR45\cite{2017Remote} can achieve a slight improvement over knowledge transfer learning from ImageNet\cite{ILSVRC15} to AID\cite{2017AID} or NR45\cite{2017Remote}. Except for the scene classification task, related to land cover classification in RSD, X. Tong \textit{et al.}\cite{tong2020land} created a land cover classification benchmark called the Gaofen Image Dataset (GID) which consists of 150 high-resolution Gaofen-2 (GF-2) satellite images with 5 classes of pixel-level annotation and 10 images with finer pixel-level annotation of 15 categories. In \cite{tong2020land}, the knowledge transfer learning method was also used for land cover classification tasks. First, ResNet-50\cite{he2016deep} is pretrained by 150,000 image patches of GID \cite{tong2020land} to facilitate the generalization ability of pixel-wise land cover classification, and then, based on the pretrained ResNet-50\cite{he2016deep}, they also used 30,000 image patches for model fine-tuning to obtain the highest performance of pixel-wise land cover classification of 15 categories. Next, P. Ranjan \textit{et al.}\cite{ranjan2020build} also proposed a transfer learning framework for boosting multiscale pixel-wise footprint extraction from aerial images, which is based on multi-resolution data pretraining and fine-tuning. Here, \cite{ranjan2020build} utilized the subband data from the 2-D discrete wavelet transform of original images to pretrain the encoder of the Unet architecture\cite{ronneberger2015u}, and then the pretrained model was migrated into the original image fine-tuning step to achieve multi-resolution transfer learning. Y. Liu \textit{et al.} \cite{liu2018semantic} utilized the PASCAL VOC\cite{everingham2015pascal} to pretrain the encoder of their model, and then the pretrained model is fine-tuned on fewer specific land cover labeling data to provide acceptable pixel-wise prediction results. With regard to the object detection task in RSD, K. Li \textit{et al.}\cite{2019Object} set up an available benchmark for object detection in optical remote sensing images called DIOR, which contained 23,463 images and 192,472 object instances within 20 common object categories. 
    Then, \cite{2019Object} pretrained all detectors on large-scale nature scene datesets (e.g., ImageNet\cite{ILSVRC15} and Microsoft Common Object in Context (MS COCO)\cite{2014Microsoft}) and then fine-tuned them on DIOR\cite{2019Object} to achieve the knowledge transfer learning of object detection from natural scenes to remote sensing scenes. In addition, by their experimental results on DIOR\cite{2019Object}, we find that the knowledge transfer learning method also makes the superior detectors in natural scenes obtain competitive results in RSD without any careful model design. Then, related to the ship detection task from SAR images, J. Li \textit{et al.}\cite{li2017ship} proposed an improved Faster R-CNN\cite{ren2015faster} and set up an SAR Ship Detection Dataset (SSDD) as a benchmark to evaluate the proposed detector compared with Faster R-CNN\cite{ren2015faster}. In \cite{li2017ship}, due to the smaller scale and large difference imaging character of SSDD, they designed a knowledge transfer learning method to pretrain the first five convolution layers of ZFNet\cite{krizhevsky2012imagenet} on ImageNet\cite{ILSVRC15} and then fine-tune the latter two layers on SSDD dataset to obtain better ship detection performance from SAR images. Moreover, Z. Chen \textit{et al.}\cite{chen2018end} also utilized the knowledge transfer learning method to pretrain the backbone on a large-scale nature scene dataset and then fine-tuned the detectors based on the pretrained backbone, to achieve better airplane detection performance in RSD with limited training samples. In general, based on the abovementioned studies, a consensus has been basically formed. When facing data-hungry model training under data scarcity conditions, it is necessary to first generate a domain-level knowledge from large-scale datasets (e.g., ImageNet\cite{ILSVRC15}, Place365\cite{zhou2017places}, COCO\cite{2014Microsoft} and PASCAL VOC\cite{everingham2015pascal}) and then adapt to downstream tasks in RSD via fewer specific task labeling data fine-tuning.
    \begin{figure*}
    \vspace{0.2cm}
	\centering 
	\includegraphics[width=7in]{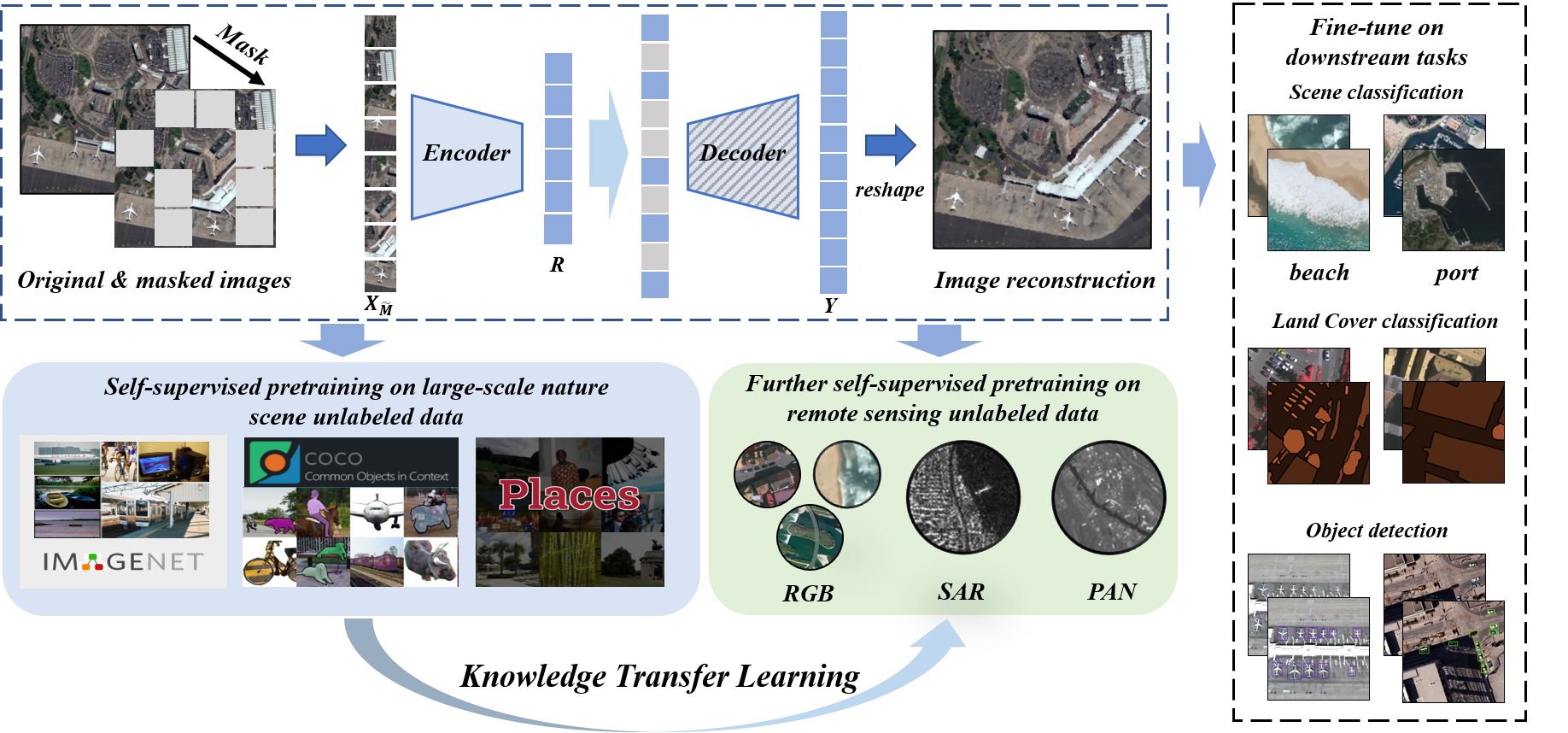}
	\caption{The overall framework of our proposed CSPT.}
	\vspace{-0.6cm}
    \end{figure*}
    \par Recently, to further explore the potential of knowledge transfer learning for task-aware model training in RSD, D. Wang \textit{et al.}\cite{wang2022empirical} provided an empirical study of model pretraining in RSD following the idea of \cite{long2022aerial}, which utilized the self-built optical remote sensing dataset M-AID\cite{long2022aerial} to pretrain models of CNNs\cite{simonyan2014very,he2016deep} and ViTs\cite{liu2021swin,zhang2022vitaev2} and then fine-tuned them on several downstream tasks to avoid the severe domain gap impact of transferring knowledge from the natural scenes to remote sensing scenes. From the experimental results of \cite{wang2022empirical}, we find that ViT-based models\cite{liu2021swin,zhang2022vitaev2} are promising backbones to provide a stronger feature representation to facilitate downstream tasks in RSD. Meanwhile, the study of \cite{wang2022empirical} also indicated that utilizing a large-scale dataset, whether belonging to the nature or remote sensing domain, for model pretraining can promote almost all downstream tasks in RSD. It seems to be the right way to construct a high quality and extreme large-scale dataset for model pretraining because it can generate domain-level knowledge and then be generalized to any domain and downstream task. Thus, to set up a transferable visual model, CLIP\cite{radford2021learning} collected a large-scale and high quality data of 400 million image-text pairs from the internet for model pretraining by multimodal supervision. Although CLIP\cite{radford2021learning} is pretrained by an extremely large-scale dataset and performs well on zero- or few-shot learning tasks in natural scenes, its experimental results pointed out that CLIP\cite{radford2021learning} underperforms on several specific, complicated or abstract tasks, especially for satellite image classification tasks. According to \cite{wang2022empirical} and \cite{radford2021learning}, under supervised learning, the current knowledge transfer learning strategy seems to be a nonoptimal solution because the severe domain gap between nature and remote sensing scenes, task-aware discrepancy from diverse downstream tasks and insufficient high-quality data annotation in RSD are nonnegligible factors that restrict large-scale pretraining and affect the performance of straightforward pretraining-then-fine-tuning on diverse downstream tasks.
    \par Consequently, to construct a more effective knowledge transfer learning method to promote task-aware model training in RSD, how to bridge the domain gap between nature and remote sensing scenes, mitigate task-aware discrepancies from diverse downstream tasks and avoid expensive manual labeling costs should be further explored and studied. In this paper, considering the task-agnostic representation of masked image modeling (MIM) applied for self-supervised pretraining and inspired by the idea of not stopping pretraining in natural language processing (NLP)\cite{gururangan2020don,dery2021should}, a concise and effective knowledge transfer learning strategy called ConSecutive PreTraining (CSPT) is designed for task-aware model training to promote almost all downstream tasks of RSD, as shown in Fig. 2. Here, the ViT-based encoder-decoder architecture is employed for model pretraining and then only using the pretrained encoder for fine-tuning on diverse downstream tasks. Different from the current knowledge transfer learning method via supervised learning, we utilize the task-agnostic representation of MIM for the consecutive self-supervised pretraining process both on natural and remote sensing scenes, which can capture the intrinsic pattern, structures, relationships and semantic by randomly masked region reconstruction to establish a more effective and robust feature representation for remote sensing scenes. Obviously, after the step of self-supervised pretraining on a large-scale nature scene dataset and before fine-tuning on specific downstream tasks, the further self-supervised pretraining on RSD can bridge the domain gap and elegantly transfer the domain-level knowledge of natural scenes to remote sensing scenes. Meanwhile, since self-supervised pretraining does not require extra annotations, a large amount of relevant unlabeled data can be joined into a consecutive self-supervised pretraining process, which can not only leverage the domain-level knowledge generated from natural scenes but also greatly release the huge potential of unlabeled data for task-aware model training in RSD. 
    \par Finally, extensive experiments are performed on a large-scale nature scene dataset (i.e., ImageNet\cite{ILSVRC15}) and twelve remote sensing datasets (i.e., AID\cite{2017AID}, NR45\cite{2017Remote}, ISPRS Potsdam and Vaihingen\cite{isprs}, GID\cite{tong2020land}, DIOR\cite{2019Object}, NWPUVHR-10\cite{cheng2014multi}, UCAS-AOD\cite{2015Orientation}, HRSC2016\cite{liu2017high}, MSTAR\cite{1998Moving}, HRSID\cite{wei2020hrsid} and SSDD\cite{li2017ship}). These datasets involve three downstream tasks (i.e., scene classification, object detection and land cover classification) and two categories of imaging data (i.e., optical RGB images and SAR). From the experimental results, we find that when more task-relevant unlabeled data is joined into the further self-supervised pretraining step or waiting for more iterative epochs, the newly designed CSPT can achieve the promising performance even reaching the state-of-the-art (SOTA) result without any expensive labeling consumption and careful model design. In summary, the contributions of our study can be summarized below:
\begin{enumerate}
\setlength{\parskip}{0.1cm}
    \item For the current knowledge transfer learning strategy based on supervised pretraining-then-fine-tuning from the natural scene domain to the RSD, the inevitable annotation noise and severe domain gap are analyzed in detail. In addition, a concise and effective knowledge transfer learning strategy called CSPT is proposed to gradually bridge the domain gap, so that it can efficiently transfer the domain-level knowledge of the nature scene domain into RSD. Meanwhile, the designed CSPT can be a promising way to release the huge potential of unlabeled data for task-aware model pretraining. 
    \item Based on the task-agnostic representation of MIM as a pretext task, we study the impact of adding extra relevant unlabeled data or waiting for more iterative epochs for further self-supervised pretraining in RSD. We find that the designed CSPT can be a more feasible way to promote the fine-tuning performance than manually annotating a large-scale and high-quality dataset for task-aware model training in RSD.
    \item Extensive experiments are conducted, including three downstream tasks of scene classification, object detection and land cover classification in RSD. The experimental results show that the designed CSPT can mitigate task-aware discrepancy and advance the performance of diverse downstream tasks to reach competitive results in comparison with SOTA methods. Finally, we make the pretrained model weights freely available at \href{https://github.com/ZhAnGToNG1/transfer\_learning\_cspt}{https://github.com/ZhAnGToNG1/transfer\_learning\_cspt} to the remote sensing community.  
 \end{enumerate}
\par The rest of this paper is organized as follows: Section II introduces the related works on knowledge transfer learning and self-supervised pretraining. Section III elaborates the problem analysis and newly designed knowledge transfer learning called CSPT in RSD. Extensive experiments are reported in Section IV with detailed discussions, and the conclusion is provided in Section V.
\section{Related work}
\subsection{Knowledge Transfer learning}
    Knowledge transfer learning is a fundamental and important issue in deep learning. It is widely used for task-aware model training to prevent overfitting when it is difficult to obtain sufficient data for downstream tasks. In computer vision field, some large-scale nature scene datasets, such as ImageNet\cite{ILSVRC15}, Place365\cite{zhou2017places} or COCO\cite{2014Microsoft} are usually employed for data-hungry model (e.g., CNNs and ViTs) pretraining. This can set up domain-level knowledge with a more generalized feature representation. Many studies\cite{2017AID,2017Remote,tong2020land,2019Object,cheng2014multi,2015Orientation,liu2017high,wei2020hrsid,li2017ship,long2022aerial,ranjan2020build,liu2018semantic,chen2018end,wang2022empirical,radford2021learning,chakraborty2020efficient,ericsson2021well,kotar2021contrasting,asano2021pass,stojnic2021self,li2021geographical,manas2021seasonal,reed2022self} have demonstrated that the set up domain-level knowledge can be easily generalized into any domain or downstream task by a fine-tuning step. Subsequently, except for knowledge transfer learning via supervised learning method \cite{2017AID,2017Remote,tong2020land,2019Object,cheng2014multi,2015Orientation,liu2017high,wei2020hrsid,li2017ship,long2022aerial,ranjan2020build,liu2018semantic,chen2018end,wang2022empirical,radford2021learning} in RSD, \cite{chakraborty2020efficient,ericsson2021well,kotar2021contrasting} began to explore self-supervised pretraining for knowledge transfer learning to relieve the burden of carefully labeling a large-scale dataset. They also pointed out that self-supervised pretraining not only can reduce the data annotation but can also generate more superior feature representation than supervised pretraining method. For example, \cite{asano2021pass} set up an unlabeled dataset called Picture without humAns for Self-Supervision (PASS) for self-supervised pretraining. As a result, it can yield a knowledge transfer learning performance on PASS that is similar to supervised pretraining on ImageNet\cite{ILSVRC15} and can adapt to downstream tasks of human pose estimation. Next, according to studies reported in \cite{chakraborty2020efficient,ericsson2021well,kotar2021contrasting,asano2021pass}, the self-supervised pretraining for knowledge transfer learning exhibits two advantages: (1) no need for expensive human labeling and (2) more generalized feature representation, which are friendly to downstream tasks in highly specialized areas such as RSD. Thus, in \cite{stojnic2021self}, the authors analyzed the applicability of self-supervised pretraining for knowledge transfer learning with different numbers and domains of unlabeled images adapting to scene classification task in RSD. By analyzing the results, they indicated that the self-supervised pretraining on unlabeled remote sensing images can provide better results than supervised pretraining on a large-scale dataset of natural scenes, even when they used significantly fewer unlabeled images. In addition, \cite{stojnic2021self} also found that their constructed self-supervised pretraining method of contrastive multiview coding can be easily extended to knowledge transfer learning from unlabeled multispectral images to downstream tasks in RSD. Next, some studies \cite{li2021geographical,manas2021seasonal} aimed to fully utilize unlabeled images in RSD and avoid the severe domain gap between nature and remote sensing scenes. Thus, they set up transferable representations by collecting unlabeled remote sensing images with geographical and seasonal information for self-supervised pretraining. The results showed that they both obtain better knowledge transfer learning performance than supervised pretraining on ImageNet\cite{ILSVRC15} while adapting for multiple downstream tasks in RSD. Furthermore, \cite{reed2022self} also studied hierarchical self-supervised pretraining for knowledge transfer learning. They found that self-supervised pretraining on ImageNet\cite{ILSVRC15} can improve the self-supervised pretraining on RSD. It can also reduce the convergence time and perform better on diverse downstream tasks than in-domain self-supervised pretraining carried out from scratch. In general, self-supervised pretraining-based knowledge transfer learning is a promising way to promote task-aware model training in RSD. However, the abovementioned studies can only achieve limited performance improvements over previously supervised pretraining methods. Accordingly, in our work, we would like to study a more effective and uniform knowledge transfer learning strategy based on self-supervised pretraining in RSD to bridge the domain gap between nature and remote sensing scenes and mitigate the task-specific discrepancies in diverse downstream tasks.
\subsection{Self-Supervised Pretraining}
    \par Self-supervised pretraining is an unsupervised learning method. It is usually used to capture intrinsic patterns and semantic representations from original data. Until now, it has been a good way to significantly boost performance of downstream tasks by learning a universal and transferable representation without any expensive labeling. According the investigation and experimental analysis, the studies of \cite{anand2021recent} and \cite{ericsson2021self} have demonstrated that adopting self-supervised pretraining to learn a transferable representation tightly relies on three elements: (1) the amount and domain of training sample collection; (2) the pretext task setting; and (3) the network architecture chosen. Consequently, \cite{tao2022tov} carefully designed a sample collection strategy to automatically capture unlabeled samples with class-balanced resampling both in natural and remote sensing scenes. Then they employed the pretext task of instance-level discrimination by contrastive learning, which makes the different augmented views (i.e., positive sample pairs) of the same images closer and separates views (i.e., negative sample pairs) of different images. Subsequently, \cite{tao2022tov} showed that by utilizing their proposed self-supervised pretraining method, universal and transferable feature representation can be obtained and adapted to tasks of scene classification, object detection and semantic segmentation in RSD. Next, for different pretext task settings, \cite{xu2021adversarial} designed a novel unsupervised adversarial contrastive learning way to pretrain a CNN-based Siamese network by a pretext task of invariant representation learning, which minimized the feature similarity of augmented data and its corresponding unsupervised adversarial samples. Through the designed self-supervised pretraining method, \cite{xu2021adversarial} obtained competitive classification results on SAR target recognition datasets. In addition, to use prior information to guide self-supervised pretraining, \cite{ayush2021geography} introduced the geography-aware into the pretext task of invariant representation learning which makes the positive pairs closer than typical unrelated negative pairs and meanwhile predicts the geo-location information of input images. When unlabeled remote sensing images with geo-location prior information are applied for the self-supervised pretraining method of \cite{ayush2021geography}, a universal and transferable representation can be generated and adapted to diverse downstream tasks in RSD. Moreover, considering the task-aware discrepancy in diverse downstream tasks, \cite{li2022global} designed the global-local structure of a self-supervised pretrained CNN, which can take into account global and local pattern information from positive and negative pairs, and then the pretrained model can be transferred well into land cover classification tasks. Related to the domain of training sample collection, \cite{heidler2021self} collected the extra domain samples of audio recordings and combined them with remote sensing images to set up audiovisual representation, which has to identify the embedding feature pairs of audio and remote sensing images by the pretext task of invariant representation learning. Then, \cite{tao2020remote} also demonstrated that when collecting sufficient unlabeled data and narrowing the domain difference between collected pretraining data and task-aware fine-tuning data, a self-supervised pretraining model would perform well on downstream tasks. In addition, the correlation between pretext tasks and downstream tasks is also important for mitigating task-aware discrepancy. 
    \begin{figure*}
    \vspace{0.2cm}
	    \centering 
        \includegraphics[width=7in]{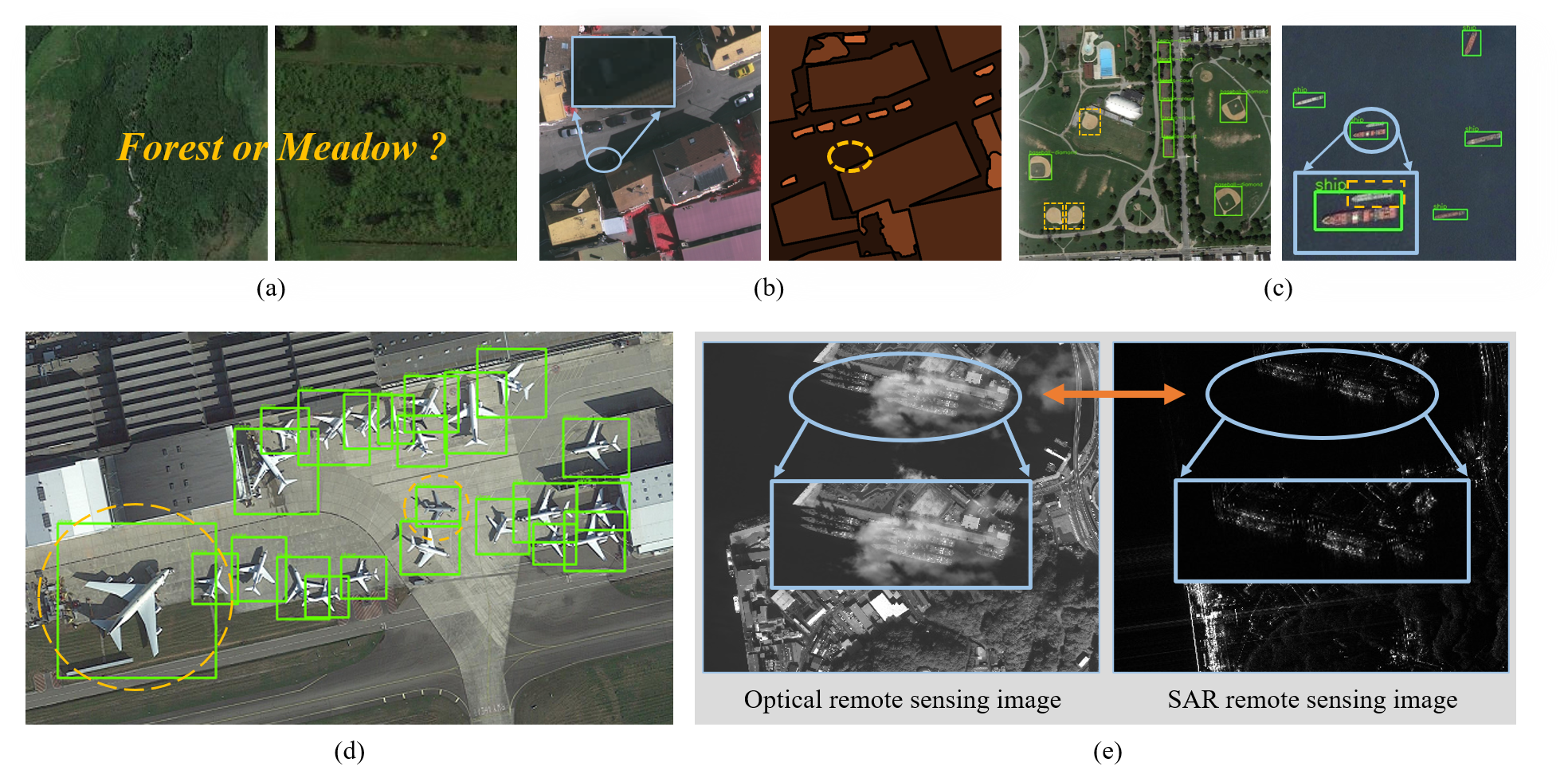}
	    \caption{The examples about various labeling problems on different tasks. (a) ambiguous definition on similar scenes; (b) the omission of hidden objects; (c) the situation of missing annotation; (d) inconsistent annotation standard; (e) the difficult annotation on the special imaging mode.}	
    \vspace{-0.6cm}
    \end{figure*}
    \par In general, the abovementioned self-supervised pretraining studies \cite{tao2022tov,xu2021adversarial,ayush2021geography,li2022global,heidler2021self,tao2020remote} are mainly based on contrastive learning such as SimCLR \cite{chen2020big}, MoCo \cite{he2020momentum}, BYOL \cite{grill2020bootstrap} and SwAV \cite{caron2020unsupervised}. Although these studies avoid expensive labeling consumption and obtain a more universal and transferable feature representation in RSD, positive and negative pairs of unlabeled data still need to be carefully set up for model pretraining. Furthermore, the global decision information from the deep layer of CNNs is often employed for pretext tasks of instance level discrimination and invariant representation. This works well for global decision task such as scene classification but not so well for dense prediction tasks (e.g., object detection and segmentation tasks) due to the task-aware discrepancy between the pretext and downstream tasks. Recently, ViTs have been shown to be more powerful network architecture than CNNs. Its patch structure coincides the idea of the pretext task of masked language modeling for self-supervised learning in NLP. Thus, some works, such as MAE\cite{he2021masked}, BEiT\cite{bao2021beit}, SimMIM\cite{xie2021simmim} have begun to turn to mask partial image patches and then reconstruct the original image (i.e., MIM task) by an encoder-decoder structure to achieve self-supervised pretraining. Among them, the studies of \cite{he2021masked} and \cite{xie2021simmim} demonstrated that the simple pretext task based on MIM (i.e., predicting the raw pixels of RGB image values) for self-supervised pretraining can provide a strong transferable ability to diverse downstream tasks than previously applied pretext tasks. In addition, the study of \cite{wang2022repre} also has demonstrated that an extra branch of task-agnostic representation of MIM in parallel with the existing contrastive learning can facilitate the self-supervised pretraining and then mitigate task-aware discrepancy from diverse downstream tasks. Following these advanced self-supervised pretraining works that used MIM-based pretext task in natural scenes, Zhou \textit{et al.} \cite{zhou2022self} utilized MAE \cite{he2021masked} for a self-supervised pretraining study on medical image analysis and showed that when using the MIM-based pretext task, the pretrained model can significantly improve the fine-tuning performance of diverse medical downstream tasks. Thus, referring to the ViT \cite{dosovitskiy2020image} architecture and task-agnostic representation of MIM, we would like to further explore a more effective and uniform self-supervised pretraining method in RSD. While doing so, we also focus on analyzing the impact of the amount and domain of training data collection, network architecture selection and different pretext task settings.
\section{Knowledge Transfer Learning Strategy}
    In this section, the problems of label noise and domain gap for knowledge transfer learning are first analyzed in Section \uppercase\expandafter{\romannumeral3}-A. Next, in Section \uppercase\expandafter{\romannumeral3}-B, we introduce the proposed knowledge transfer learning method in detail. Finally, the mechanism of the task-agnostic representation of MIM is revisited in Section \uppercase\expandafter{\romannumeral3}-C.
\subsection{Problem Analysis}
    \textbf{\emph{Label Noise:}} Recently, in RSD, several works \cite{long2022aerial,wang2022empirical} have considered the task-aware model training via supervised pretraining on a self-built large-scale dataset and then fine-tuning on downstream tasks. However, setting up a large-scale and well-annotated dataset from RSD involves expensive labor. In addition, considerable label noise is inevitably generated during manual annotation. This affects the supervised pretraining and fine-tuning steps. In Fig. 3, we enumerate some frequently occurring annotation issues. For instance, in Fig. 3 (a), in scene classification, due to very similar texture, structure and content observed from overlooking view, these two scenes from AID \cite{2017AID} can be easily confused and lead to manual labeling as "forest" or "meadow". Next,  land cover classification not only has to annotate lots of pixels but also needs to identify the category of each pixel. This easily results in some categories of pixel annotations being omitted under complex semantic relations, such as the small vehicle being hidden in the shadow of buildings, as in the ISPRS dataset \cite{isprs} and shown in Fig. 3 (b). Then, for object detection annotation in optical remote sensing scenes, finding several valid objects from large-scale optical remote sensing scenes is very difficult. Thus, missing labeling situations often occur. For example, in Fig. 3 (c), the baseball diamond and densely arranged ships are not annotated in the published dataset (e.g., NWPUVHR-10\cite{cheng2014multi}). Next, due to subjective inconsistency, it is difficult to form the unified standard for manually labelling bounding boxes, which is illustrated in Fig. 3 (d). The yellow dashed circles of airplane annotations indicate that two bounding boxes are annotated inconsistency (e.g., the left box contains more background information than the right box in DOTA \cite{xia2018dota} dataset). Furthermore, for labeling different types of imaging data, as illustrated in Fig. 3 (e), if there is not corresponding optical remote sensing scene image as a reference, it is difficult to annotate ships on SAR images. In general, the wrong label information for scene classification, insufficient and incorrect pixel-level annotations for land cover classification, defective and inconsistent object-level annotations for object detection and lack of professional knowledge for different types of imaging data annotation in RSD, can all lead to label noise. This will directly affect the model convergence and guide it to learn an inaccurate supervised signal according to a self-built large-scale dataset. As a result, it could have a negative impact in the pretraining or fine-tuning step and limit the downstream task performance. Consequently, in RSD, it is worthy studying how to abandon the dependency of large-scale data annotation and release the huge potential of unlabeled data for task-aware model training.  
    \par \textbf{\emph{Domain Gap:}} As mentioned in Section \uppercase\expandafter{\romannumeral1} and \uppercase\expandafter{\romannumeral2}, the domain gap is a troublesome issue in knowledge transfer learning which limit the performance of task-aware model training. At present, many works have shown that self-supervised pretraining is a promising way to learn a transferable representation than supervised pretraining. In particular, MAE \cite{he2021masked} has been utilized for self-supervised pretraining on the unlabeled nature scene dataset of ImageNet \cite{ILSVRC15} based on the ViT \cite{dosovitskiy2020image} architecture. For intuitive analysis, the pretrained weights of ViT-B \cite{dosovitskiy2020image} from MAE \cite{he2021masked} are adopted to individually generate visualized attention scores from two unseen images. They are then used to indicate the existing domain gap between nature and remote sensing scenes. The visualized attention scores are calculated from the last self-attention layer of ViT-B \cite{dosovitskiy2020image} via the query-key product, and the high attention score is the red color area in self-attention map. In Fig. 4 (a), refer to the selected red rectangle area located in a vehicle of nature scene image, high attention scores reveal that the self-supervised pretraining weights of ViT-B \cite{dosovitskiy2020image} can correctly pay attention to relevant areas of the vehicle. However, when the same pretrained weights of ViT-B\cite{dosovitskiy2020image} are directly applied to remote sensing image, as shown Fig. 4 (b), some high attention scores are distributed in areas irrelevant to the previously selected red rectangle area, and only a few high attention scores pay attention to relevant areas of vehicles in the remote sensing image. Therefore, from the visualized results of Fig. 4 (a) and (b), we can see that the domain gap does exist between nature and remote sensing scenes because the weights of ViT-B\cite{dosovitskiy2020image} self-supervised pretrained on unlabeled nature scene data can be more aware of unseen nature scenes than unseen remote sensing scenes. Subjectively, in Fig. 4 (a) and (b), except for the difference in appearance of the vehicle in nature and remote sensing scenes, the context information of the vehicles is also different. For example, the vehicle in natural scene has more fixed context information because wheels are always on the ground and the top of the vehicle is toward the sky; however, vehicles in remote sensing scenes would have more flexible context information because they can appear in any area with complex and various surroundings caused by the overlooking view.
    \par Considering the domain gap between nature and remote sensing scenes, we also quantitatively analyze the impact of the domain gap for the fine-tuning step in different knowledge transfer learning methods. Specifically, there are three different curves of fine-tuning loss presented in Fig. 4 (c) and (d), where, the x-axis represents the epoch, and the y-axis represents the corresponding fine-tuning loss value. In addition, the different color curves represent three different pretraining methods: (1) the blue curve represents that the model is self-supervised pretrained on unlabeled ImageNet-1K\cite{ILSVRC15} called SSP (IN1K); (2) the orange curve indicates that the model is supervised pretrained on a ready-made ImageNet-1K\cite{ILSVRC15} called SP (IN1K); (3) the green curve means that the model is self-supervised pretrained on unlabeled ImageNet-1K\cite{ILSVRC15} and then further self-supervised pretrained on the training data of AID\cite{2017AID} and NR45\cite{2017Remote} called SSP (IN1K$\rightarrow$Train). Notably, the above pretraining methods all adopt ViT-B\cite{dosovitskiy2020image}. Subsequently, all three different pretrained models from SSP (IN1K), SP (IN1K) and SSP (IN1K$\rightarrow$Train) are fine-tuned by AID\cite{2017AID} and NR45\cite{2017Remote}. From Fig. 4 (c) and (d), we can see that the blue curve of SSP (IN1K) can converge to a lower-level loss value at the fine-tuning step than the orange curve of SP (IN1K). This indicates that the self-supervised pretraining method can indeed generate more transferable representation than the supervised pretraining method to leverage the fine-tuning step in knowledge transfer learning. Furthermore, because the self-supervised pretraining applied to easily collected large-scale unlabeled nature scene data can generate domain-level knowledge, the further self-supervised pretraining process on the unlabeled remote sensing dataset is considered to adapt the generalized domain-level knowledge into RSD. From the green curves of SSP (IN1K$\rightarrow$Train), we can see that whether at the initial or the end of state of fine-tuning step, the fine-tuning loss values of SSP (IN1K$\rightarrow$Train) are both lower than those of SSP (IN1K) and SP (IN1K) on AID\cite{2017AID} and NR45\cite{2017Remote}. Thus, SSP (IN1K$\rightarrow$Train) can be a promising way to facilitate the fine-tuning step and gradually bridge the domain gap between nature and remote sensing scenes. From the above intuitive and quantitative analysis, the domain gap is an important problem that cannot be ignored for task-aware model training in RSD.

    \begin{figure*}
    \vspace{0.2cm}
	\centering 
    \includegraphics[width=7in]{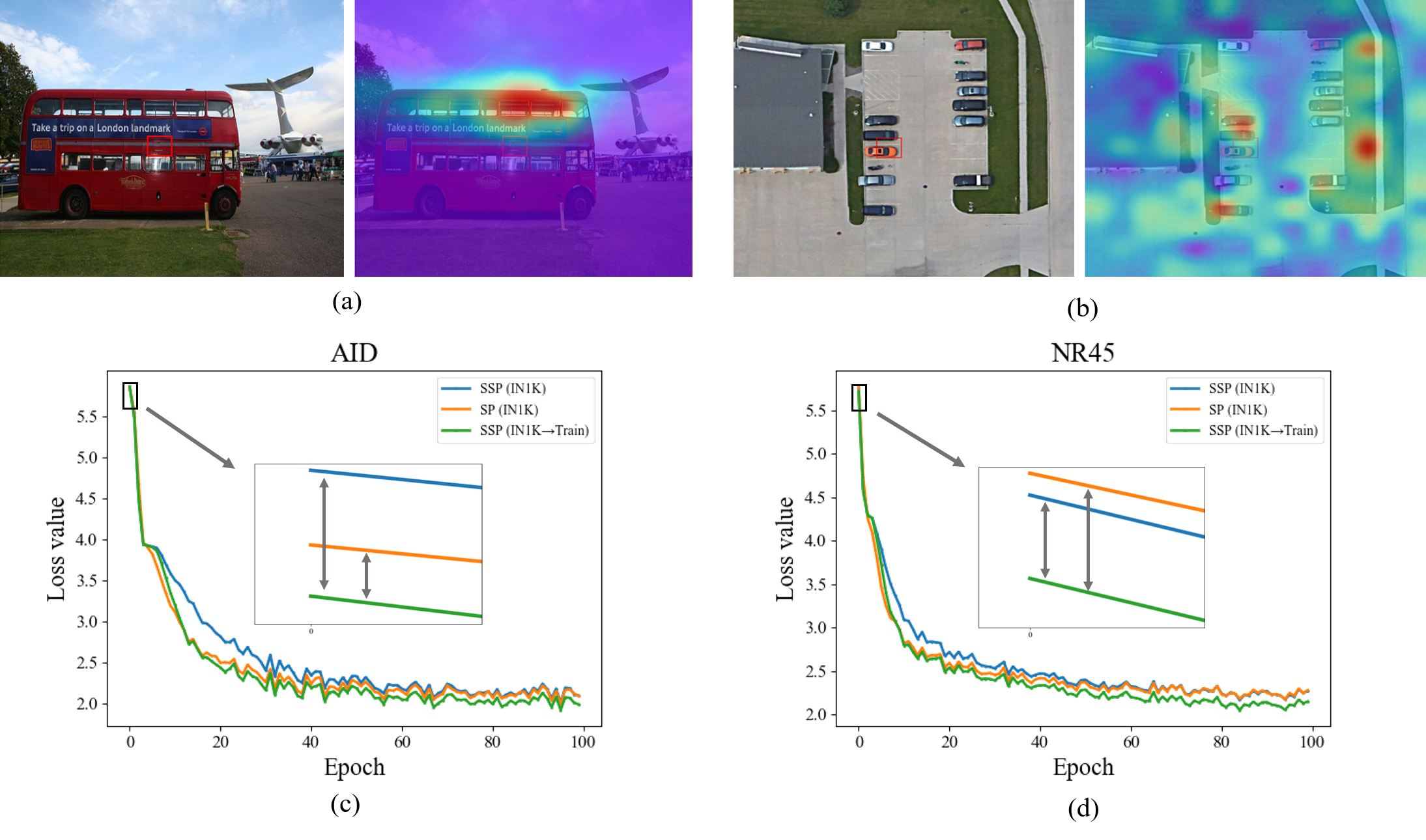}
	\caption{The intuitive and quantitative analysis on domain gap between natural scene and RSD. (a) is the self-attention map for a vehicle of natural scene image; (b) is the self-attention map for the vehicles of remote sensing scene image; (c) and (d) represent the fine-tuning loss convergence curves of different pretraining methods on AID\cite{2017AID} and NR45\cite{2017Remote}.}	
    \vspace{-0.6cm}
    \end{figure*}

\subsection{CSPT for Knowledge Transfer Learning} 
    \par Taking into account the problem analysis in Section \uppercase\expandafter{\romannumeral3} A and inspired by \cite{gururangan2020don,dery2021should} in NLP, a concise and effective knowledge transfer learning strategy called CSPT is proposed for task-aware model training in RSD. As shown in Fig. 2, the overall framework is composed of three steps: (1) self-supervised pretraining on a large-scale unlabeled nature scene data; (2) further self-supervised pretraining on task-related unlabeled remote sensing data; and (3) fine-tuning on diverse downstream tasks. Here, following the MAE \cite{he2021masked}, we adopt the MIM-based pretext task for self-supervised pretraining both in steps (1) and (2). Similarly, the benefits of not stopping pretraining model have been successfully verified in NLP\cite{gururangan2020don}. The pretext task of masked/auto-regressive language modeling (MLM) is employed for transformer-based model pretraining via masked and then completed tokens of words in each sentence. This enable the pretrained language model to learn the vocabulary, sentence structure, semantic and even understand the context of large-scale unlabeled text data. First, a generalist language model such as BERT \cite{devlin2018bert} or GPT \cite{gpt} is firstly pretrained by large-scale unlabeled text data to set up domain-level knowledge. Then, continuously pretraining the generalist model in NLP can make the set up domain-level knowledge easily adapt to any domain (e.g., biomedical, computer science publications, news, or reviewers) or downstream task by domain adaptive pretraining and fine-tuning steps. 
    \par Along with the transformer structure of NLP being migrated into computer vision filed, ViT\cite{dosovitskiy2020image} set up vision words (e.g., a group of pixels such as \(16\times16\) patches) from 2-D images, which provides an opportunity to pretrain a model like the way of NLP. Moreover, some pioneering studies\cite{bao2021beit,el2021large,he2021masked,xie2021simmim,maskfeat} have also demonstrated the superiority of MIM-based pretext task applied for self-supervised pretraining on large-scale unlabeled natural scene images. Notably, the MIM-based pretext task is similar to MLM. If ViT-based model can reconstruct randomly masked partial tokens of vision words, it indicates that the model has learned the pattern relation, structure, semantic and even understood the content of unlabeled images. Thus, as shown in step (1) of Fig. 2, when the ViT-based model pretraining on a large enough unlabeled dataset such as ImageNet\cite{ILSVRC15} by self-supervised learning, the various combinations of pattern relation, structure, semantic and even content can be familiarized to set up domain-level knowledge and prepared for adapting to RSD. The domain-level knowledge that has a powerful feature representation and transferable ability. However, as discussed in Section \uppercase\expandafter{\romannumeral3} A, directly transferring the domain-level knowledge of the nature scene into the RSD by a fine-tuning step is a nonoptimal solution because of the severe domain gap. Therefore, we assume that the basic characters (e.g., various combinations of pixel-level pattern relation and structure) of an image are domain-invariant schema. Then, based on the generalist model with domain-level knowledge, we let it to further become familiar with the semantic and content of task-related unlabeled remote sensing images in RSD. This feel is a more reasonable method as discussed in Section \uppercase\expandafter{\romannumeral3} A. Consequently, in Fig. 2, the step (2) of further self-supervised pretraining on unlabeled remote sensing data is necessary for knowledge transfer learning. Driven by the abundant captured combinations of pattern relations and structures from large-scale nature scene data, further self-supervised pretraining generalist model of step (1) in RSD can become quickly familiar with the semantic and content of unlabeled remote sensing scene images by joining more relevant unlabeled data and waiting for more iterative epochs. As a result, the domain-level knowledge of the generalist model can be easily transferred into RSD at the further self-supervised pretraing step. Finally, in the step (3), the pretrained model allows various downstream tasks with limited labeling data still obtain good performance.
\vspace{-0.1cm}
    \begin{figure*}
    \vspace{0.2cm}
	\centering 
	\includegraphics[width=7in]{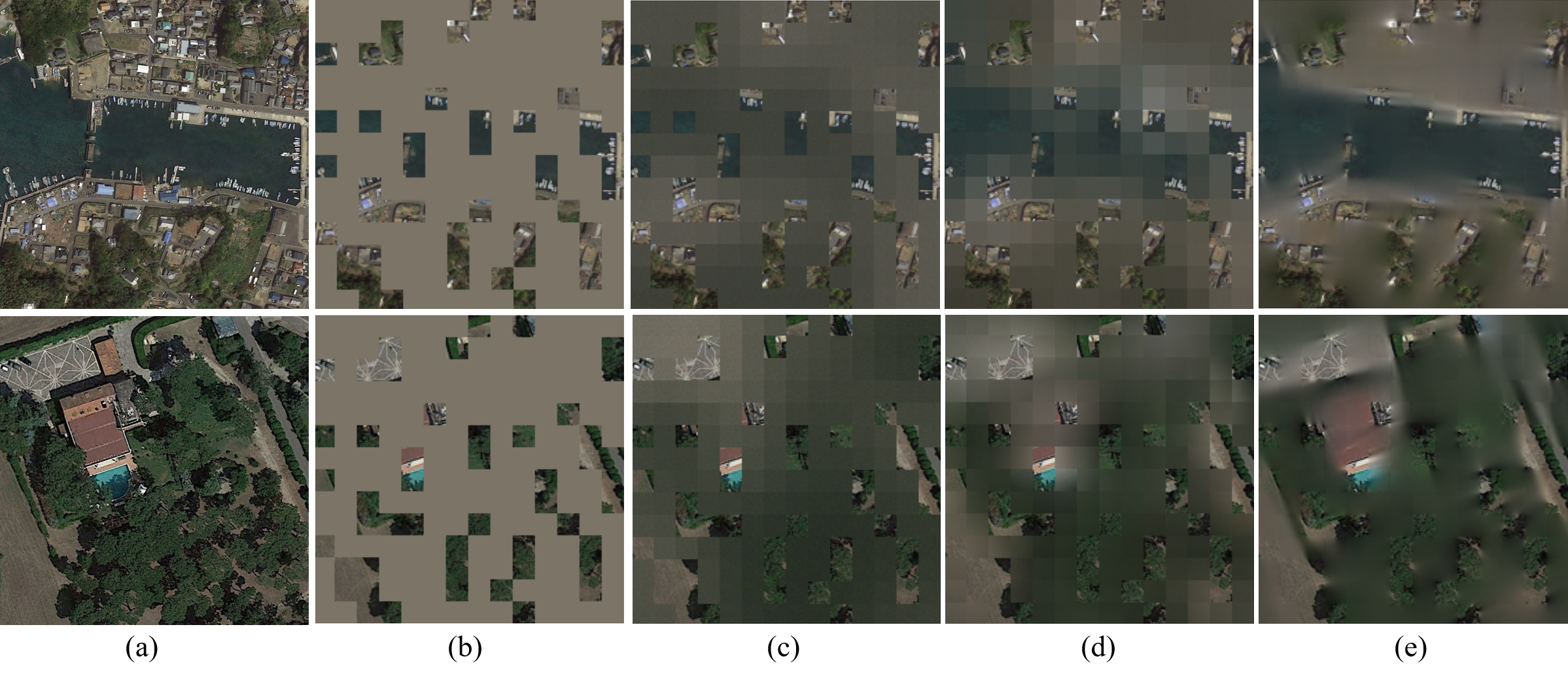}
	\caption{Reconstructed examples on AID\cite{2017AID} images. (a) denotes the original images; (b) denotes masked images; (c) denotes reconstructed results of pretrained model at 10th epoch; (d) denotes reconstructed results of pretrained model at 50th epoch; (e) denotes reconstructed results of pretrained model at 800th epoch. Specifically, the reconstruction quality from (c) to (e) gets better and better.}
    \vspace{-0.6cm}
    \end{figure*}
\subsection{Revisiting Masked Image Modeling}
    As mentioned before, the task-agnostic representation of MIM is a very effective pretext task for the designed CSPT. It randomly masks partial tokens of vision words and then reconstructs them to predict pixel values compared with their corresponding ground truth. As shown in Fig. 2, the ViT-based encoder and a lightweight transformer decoder are utilized for MIM pretext task. Next, through randomly masked region reconstruction by the decoder as illustrated in Fig. 5, the encoder can learn a powerful feature representation when the decoder can reconstruct clearer images as shown in Fig. 5 (c), (d) and (e) from masked original images (i.e., in Fig. 5 (b)). Based on the encoder-decoder architecture, the whole reconstruction process is similar to the mechanism of gradually understanding remote sensing images by humans. Thus, the encoder-decoder architecture in Fig. 2 can reasonably reveal whether the ViT-based model learned the content about input images well. In our study, the process of task-agnostic representation of MIM follows \cite{he2021masked}, and it can be expressed by the following equations:
\[R = encoder({X_{\tilde M}})\eqno(1)\]
\[Y = decoder(R)\eqno(2)\]
\[Loss = \frac{1}{{\alpha \left( {{X_M}} \right)}}{\left\| {\left. {{Y_M} - {X_M}} \right\|} \right._1}\eqno(3)\]
    In (1), ${X}$ represents the vision words of split patches from an input image, and then aiming to capture the latent representation of $R$, ${X_{\tilde M}}$ can be encoded by \(encoder\left(  \cdot  \right)\). Here, ${M}$ is the index to indicate the masked patches of ${X}$; in contrast, ${\tilde M}$ is the index to represent the unmasked patches of ${X}$. Next, in (2), the encoded $R$ can be decoded by \(decoder\left(  \cdot  \right)\) and produce the reconstructed image ${Y}$, where  ${X}$ and $Y \in {R^{H \times W \times 3}}$. In (3), the ${L_{1}}$-loss is employed to evaluate the similarity of RGB pixel values between ${X_M}$ and ${Y_M}$. Here, $\alpha \left( {{X_{M}}} \right)$ is the number of masked pixels. When the ${L_{1}}$-loss gradually decreases, the reconstructed results become clearer, which indicates that the model has captured the basic pattern relation, structure, semantic and even understood the context of the input images. Subsequently, when the model can well understand the context of images, the ViT-based encoder can be applied for fine-tuning on diverse downstream tasks to prevent task-aware discrepancy.  
\section{Experiments and Analysis}
    In this section, extensive experiments were carried out to explore the impact of the designed CSPT on diverse downstream tasks in RSD. Specifically, through these experiments, we establish three points: (1) \emph{Effectiveness:} the further self-supervised pretraining step is a necessary step for consecutive pretraining to gradually bridge the domain gap between nature and remote sensing scenes. To achieve this, it only needs to collect more task-related unlabeled data and wait for more iterative epochs. (2) \emph{Robustness:} the task-agnostic representation of MIM is applied for self-supervised pretraining steps can mitigate task-aware discrepancy while performing well on diverse downstream tasks in RSD. (3) \emph{Scalability:} when getting rid of a large-scale and carefully data annotation constraint, how feasible it is to use different volumes of unlabeled data for further self-supervised pretraining, and how data volume can impact the performance of the fine-tuning step are discussed in detail. In addition, more difficult and special SAR data in RSD is also introduced for scalability discussion and verification of the knwoledge transfer learning ability of the designed CSPT. Finally, we also compared the results with other advanced model pretraining technologies while some state-of-the-art (SOTA) methods are selected for comparison. It aims to exhibit the excellent performance of designed CSPT in RSD.
\vspace{-0.1cm}
\subsection{Datasets Description}
    To prove the effectiveness of knowledge transfer learning strategy, we adopt twelve public remote sensing datasets, including nine optical remote sensing datasets and three SAR remote sensing datasets. These datasets involve three basic downstream tasks: scene classification, object detection and land cover classification. Further details about the dataset split, category number and data volume are illustrated in Table \uppercase\expandafter{\romannumeral1}. In particular, because the image sizes of some datasets are irregular, such as ISPRS Potsdam and Vaihingen \cite{isprs}, GID \cite{tong2020land}, NWPUVHR-10 \cite{cheng2014multi} and UCAS-AOD \cite{2015Orientation}, we cropped them into 512$\times$512 subimages. Specifically, the amount of data in Table \uppercase\expandafter{\romannumeral1} is listed after cropping. Then, it can be found that the amount of remote sensing datasets is extremely limited compared with natural scene datasets. This allows us to reasonably verify the effect of knowledge transfer learning for task-aware model training in RSD.
\vspace{-0.1cm}
\subsection{Implementation Details}
    As mentioned in Section \uppercase\expandafter{\romannumeral3} B, our proposed CSPT involves three steps: two steps of self-supervised pretraining and one step of fine-tuning. Next, we would elaborate the pretraining and fine-tuning settings for all model training strategies.
\subsubsection{Pretraining Setting}
    In our work, a encoder-decoder architecture is adopted to achieve MIM-based pretext task of the task-agnostic representation. Specifically, ViT-B/L \cite{dosovitskiy2020image} is chosen as the encoder, and a lightweight transformer (e.g., depth=8) is adopted as the decoder. Then, all input images are regulated into 224$\times$224 and divided into nonoverlapping 16$\times$16 patches (i.e., vision words) as tokens. Subsequently, to ensure effectiveness, we keep the same setting of MAE \cite{he2021masked} by utilizing 75\% tokens randomly masked and then reconstructed by the decoder. In addition, the original pixel values of masked tokens are regarded as supervised signals to achieve the self-supervised pretraining. In practice, according to the first self-supervised pretraining step, the set up ViT-based model is pretrained on a large-scale unlabeled dataset of ImageNet-1K (IN1K)\cite{ILSVRC15} for 800 epochs, which can obtain a generalist model. Second, related to core idea of consecutive pretraining, the generalist model is further pretrained on relevant unlabeled data to achieve knowledge transfer learning. For experimental details, the batchsize is set to 64 on one RTX 3090. AdamW\cite{adamw} with momentum $\beta$1=0.9 and $\beta$2=0.95 is employed for optimization. The learning rate schedule adopts cosine decay with a base learning rate of 3.75e-5. Moreover, input images are augmented by random scale [0.2,1.0], random crop and random horizontal flip. For comparison, we also selected other different pretraining methods, namely, supervised pretraining on IN1K, supervised pretraining on M-AID, supervised pretraining on NR45, self-supervised pretraining on NR45 or AID, self-supervised pretraining on Sentinel-2 and self-supervised pretraining on IN1K, called SP (IN1K), SP(M-AID), SP(NR45), SSP(NR45 or AID), SSP(Sentinel-2) and SSP(IN1K).
\subsubsection{Fine-tuning Setting}
    In the fine-tuning step, we remove the decoder module of encoder-decoder architecture and apply the encoder module as the backbone network on downstream tasks. For diverse downstream tasks, different basic network architectures are chosen without careful model design. Next, the implementation details are elaborated according to different downstream tasks:
    \par \textbf{\emph{Scene classification}} To migrate the encoder of the pretrained model in scene classification, average pooling is adopted for the encoder at the last layer. Regarding the classification head, a linear classifier is added, and then the CrossEntropy loss function is used for supervised training. For experimental details, we train all classification networks for 100 epochs with a batch size of 32. The AdamW \cite{adamw} (\emph{$\beta$1}=0.9, \emph{$\beta$2}=0.999) is employed with an initial learning rate of 5e-4, and a learning rate schedule follows cosine decay. The input image size is set to 224$\times$224 spatial resolution. Augmentation technology employs AutoAugment (rand-m9-mstd0.5-inc1), label smoothing (0.1), mixup (0.8) and cutmix (1.0). For result comparison, the mean average of Top-1 classification accuracy on the test set is reported. 
    \par \textbf{\emph{Object detection}} We adopt Mask-RCNN \cite{maskrcnn_he2017} with the default setting in mmdetection framework \cite{mmdetection}. Then, the encoder of the pretrained model is regarded as the backbone of Mask-RCNN \cite{maskrcnn_he2017}, and referring to \cite{bao2021beit}, the 3rd, 5th, 7th and 11th layers of ViT \cite{dosovitskiy2020image} are selected to construct the feature pyramid network. The input image size is set as 512$\times$512, and the total number of epochs is set to 12 with a batchsize of 8. Then, momentum=0.9 and weight decay=0.0001 are adopted for SGD, and the initial learning rate is set as 0.02 and then reduced by a factor of 10 times at the 8th and 11th epochs. In addition, random flipping and random resizsing are employed for data augmentation. For result comparison, we evaluate the performance by using the mean average precision (mAP@0.5) of the PASCAL VOC object challenge \cite{everingham2015pascal}.
    \par \textbf{\emph{Land cover classification}} We make use of Upernet \cite{xiao2018upernet} within the mmsegmentation framework \cite{mmseg2020} and adopt pretrained model in the same way as the object detection task. Specifically, the input size is also set to 512$\times$512. Random cropping and random flipping are used for data augmentation. AdamW \cite{adamw} with momentum \emph{$\beta$1}=0.9, \emph{$\beta$2}=0.999 is employed for optimization. We perform fine-tuning for 96K iterations with a batch size of 2. The learning rate is set as 3e-5 with poly scheduler. Finally, we report the results by calculating mean Intersection of Union (mIoU) of all land cover categories on the test set.
\vspace{-0.1cm}
\begin{table*}[]
\vspace{0.2cm}
\centering
\caption{The comparison of different knowledge transfer strategies on remote sensing downstream tasks.\label{tab:table1}}
\renewcommand\arraystretch{1.3}
\begin{threeparttable}
\setlength{\tabcolsep}{5pt}{
\begin{tabular}{ccccccccccccc}
\toprule[1.2pt]
\multirow{4}{*}{Task}                                                                  & \multirow{4}{*}{Dataset} & \multicolumn{2}{c}{CNN}  & \multicolumn{9}{c}{Vision   Transformer}                                                                         \\ \cmidrule(r){3-4} \cmidrule(r){5-13}
                                                                                       &                           & IN1K        & M-AID      & -              & \multicolumn{2}{c}{IN1K} & \multicolumn{3}{c}{IN1K$\rightarrow$Train} & \multicolumn{3}{c}{IN1K$\rightarrow$(Train+Test)} \\ \cmidrule(r){3-4} \cmidrule(r){5-13}
                                                                                       &                           & \multicolumn{2}{c}{Sup.} & From   scratch & Sup.      & Self-sup.    & \multicolumn{3}{c}{Self-sup.}  & \multicolumn{3}{c}{Self-sup.}       \\ \cmidrule(r){3-4} \cmidrule(r){5-13}
                                                                                       &                           & -           & -          & -              & -         & ep800        & ep800    & ep1600   & ep2400   & ep800     & ep1600     & ep2400     \\ \toprule[1pt]
\multirow{2}{*}{\begin{tabular}[c]{@{}c@{}}Scene \\Classification\end{tabular}}        & NR45\cite{2017Remote}                      & 92.80       & 94.20      & 67.35          & 94.10     & 93.94        & 94.23    & 94.21    & 94.16    & \textbf{95.11}     & 94.90      & 94.84      \\
                                                                                       & AID\cite{2017AID}                       & 93.36       & 95.40      & 63.15          & 94.04     & 95.00        & 95.78    & 96.05    & 96.00    & 96.69     & 96.69      & \textbf{96.75}      \\ \hline
\multirow{3}{*}{\begin{tabular}[c]{@{}c@{}}Land Cover \\Classification\end{tabular}}   & POTSDAM\cite{isprs}                   & 76.82       & -          & 60.97          & 76.43     & 78.08        & 78.36    & 78.14    & 78.09    & \textbf{78.70}     & 77.98      & 78.19      \\
                                                                                       & VAIHINGEN\cite{isprs}                 & 72.66       & -          & 60.43          & 69.21     & 71.05        & 72.34    & 72.04    & 72.90    & \textbf{74.69}     & 74.19      & 73.07      \\
                                                                                       & GID\cite{tong2020land}                       & 51.74       & -          & 44.70          & 62.64     & 62.93        & \textbf{64.97}    & 62.82    & 63.58    &  63.31         & 64.69           & 64.55           \\ \hline
\multirow{4}{*}{\begin{tabular}[c]{@{}c@{}}Object  \\Detection\end{tabular}}           & NWPUVHR-10\cite{cheng2014multi}                & 88.60       & -          & 54.80          & 68.20     & 86.00        & 87.10    & 87.20    & 87.50    & 88.40     & 88.30      & \textbf{88.90}      \\
                                                                                       & DIOR\cite{2019Object}                      & 66.70       & -          & 36.90          & 52.70     & 66.80        & 68.30    & 68.20    & 67.60    & \textbf{69.80}     & 69.20      & 68.50      \\
                                                                                       & UCAS-AOD\cite{2015Orientation}                  & 90.10       & -          & 49.00          & 83.30     & 88.70        & 89.40    & 90.00    & 89.30    & 90.00     & 90.10      & \textbf{90.30}      \\
                                                                                       & HRSC2016\cite{liu2017high}                  & 90.00       & -          & 30.00          & 82.60     & 83.00        & 89.00    & 89.40    & 89.20    & 89.60     & 89.90      & \textbf{90.10}      \\ \toprule[1.2pt]
\end{tabular}
}
\begin{tablenotes}
    \footnotesize
    \item *Note: 1) CNN architecture adopts ResNet-101\cite{he2016deep}, and Vision Transformer adopts the ViT-B \cite{dosovitskiy2020image}. 2) Evaluation metric: mean average of Top-1 classification accuracy for scene classification; mean Intersection of Union (mIoU) for land cover classification; mean average precision (mAP@0.5) for object detection.
\end{tablenotes}
\end{threeparttable}
\vspace{-0.6cm}
\end{table*}
\vspace{-0.1cm}
\subsection{Transfer Learning Ability Comparison}
    In this section, to evaluate the effectiveness of the designed CSPT, the widely used ResNet-101 \cite{he2016deep} and recently proposed ViT-B \cite{dosovitskiy2020image} are chosen as backbone networks. Then, different pretraining methods of SP(IN1K), SP(M-AID) and SSP(IN1K) are employed as knowledge transfer learning strategies and then use the same fine-tuning setting on diverse downstream tasks. Specifically, we consider nine public optical remote sensing datasets (e.g., two scene classification datasets, namely, AID\cite{2017AID} and NR45\cite{2017Remote}; three land cover classification datasets, namely, Potsdam\cite{isprs}, Vaihingen\cite{isprs} and GID\cite{tong2020land}; and four object detection datasets, namely, NWPUVHR10\cite{cheng2014multi}, DIOR\cite{2019Object}, UCAS-AOD\cite{2015Orientation} and HRSC2016\cite{liu2017high}).
    \par The experimental results are reported in Table \uppercase\expandafter{\romannumeral2}. First, as shown in the 5th column, compared with other knowledge transfer learning methods, the task-aware model training from scratch (i.e., without knowledge transfer learning) obtains the worst performance on all downstream tasks. This illustrates that knowledge transfer learning is indeed important for task-aware model training in RSD. Second, as shown in the 3rd and 6th columns, under the current widely used model pretraining method of SP(IN1K), the different backbones of ViT-B \cite{dosovitskiy2020image} and ResNet-101 \cite{he2016deep} yield different results on downstream tasks. Due to the excellent global feature capture capability of ViT architecture, the pretrained ViT-B \cite{dosovitskiy2020image} does well on scene classification datasets of AID\cite{2017AID} and NR45\cite{2017Remote} and is better than the pretrained ResNet-101 \cite{he2016deep}. For densely predicted downstream tasks of object detection and land cover classification, the performance of pretrained ResNet-101 \cite{he2016deep} on Potsdam\cite{isprs}, Vaihingen\cite{isprs}, NWPUVHR10\cite{cheng2014multi}, DIOR\cite{2019Object}, UCAS-AOD\cite{2015Orientation} and HRSC2016\cite{liu2017high} is better than pretrained ViT-B\cite{dosovitskiy2020image}. Next, as reported in the 4th column of Table \uppercase\expandafter{\romannumeral2} and referring to the study of \cite{long2022aerial}, the newly self-built scene classification dataset of M-AID is employed for SP(M-AID) to narrow the domain gap of data between the model pretraining and fine-tuning steps. This can bring the performance improvements on scene classification datasets of AID\cite{2017AID} and NR45\cite{2017Remote}. From 3rd, 4th and 6th columns of Table \uppercase\expandafter{\romannumeral2}, one can see that although the model pretraining method of SP(M-AID) can avoid the severe domain gap between nature and remote sensing scenes and bring a slight performance improvement, it has a very expensive labor consumption with inevitably occurring label noise in a large-scale self-built dataset, which would restrict further performance improvements both at the model pretraining and fine-tuning steps. Third, as reported in the 7th column of Table \uppercase\expandafter{\romannumeral2} and compared with supervised pretraining of ViT-B\cite{dosovitskiy2020image} (i.e., SP(IN1K)) in the 6th column, when SSP(IN1K) is applied for model pretraining with 800 epochs based on ViT-B\cite{dosovitskiy2020image}, we find that the performance of almost all downstream tasks is improved. This indicates that when removing the labeling constraint, the generalist model with domain-level knowledge can be set up by SSP(IN1K) to facilitate downstream tasks. Although most current studies\cite{2017AID,2017Remote,isprs,tong2020land,2019Object,cheng2014multi,2015Orientation,liu2017high,1998Moving,wei2020hrsid,li2017ship,long2022aerial,ranjan2020build,liu2018semantic,chen2018end,wang2022empirical} suggest that domain-level knowledge can also be formed by large-scale supervised pretraining methods of SP(IN1K) and SP(M-AID), obviously, they have a specific optimization supremum and provide the relatively solidified domain-level knowledge because of the quality of manually labeling data and the binding supervision signal. Accordingly, when the model is pretrained by SP(IN1K) or SP(M-AID) and then directly fine-tuned on diverse downstream tasks, some obstacles such as label noise, domain gap and task-aware discrepancy, restrict the performance improvement. Thanks to the task-agnostic representation of MIM applied to self-supervised pretraining, SSP(IN1K) captures the intrinsic pattern relationships, structures and semantic to construct domain-level knowledge with stronger plasticity. Hence it is more easily adapted to diverse downstream tasks by a fine-tuning step than SP(IN1K) and SP(M-AID).
\begin{figure}
\vspace{0.2cm}
	\centering 
	\includegraphics[width=3in]{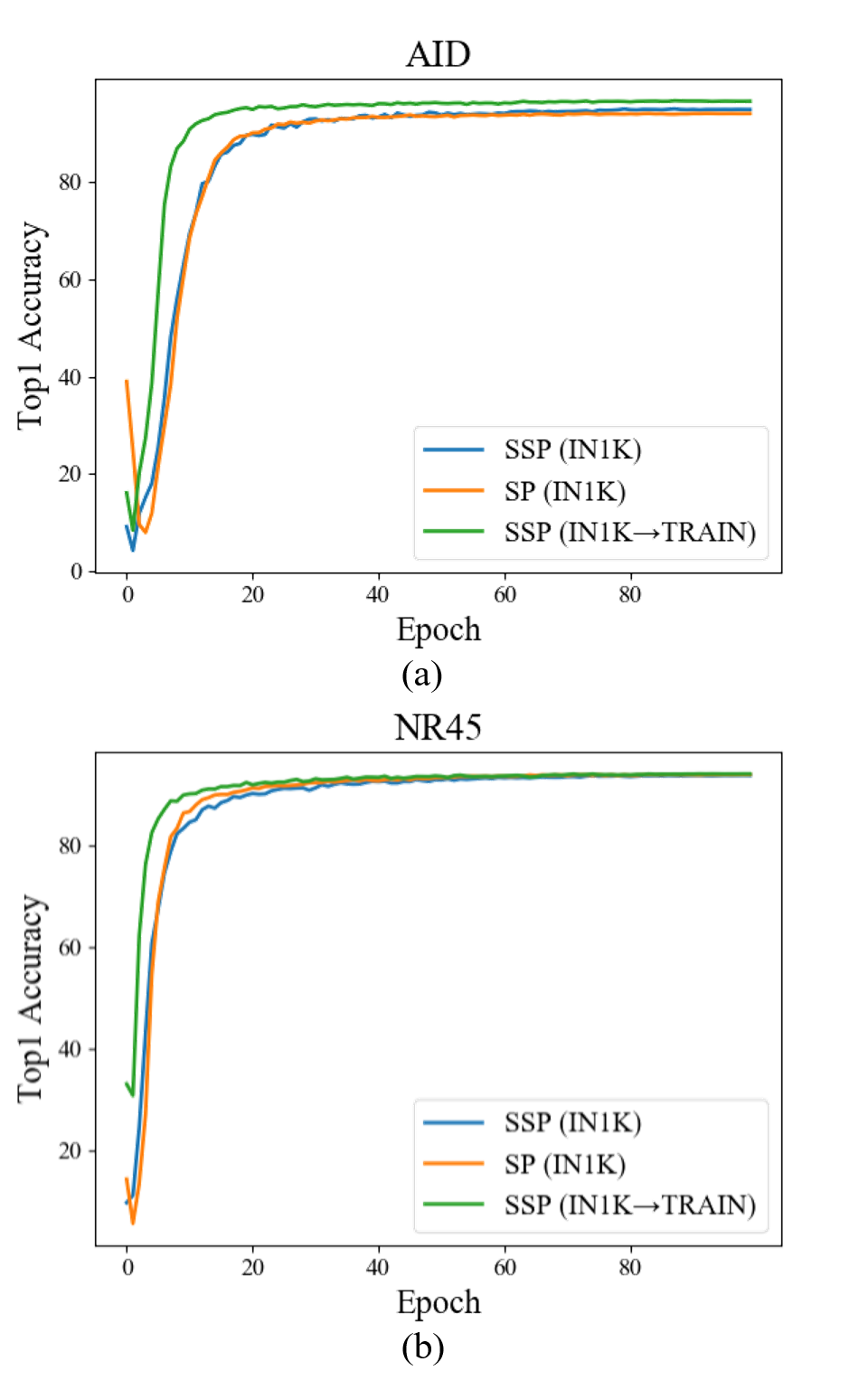}
	\caption{The Top1. accuracy curves of three different pretraining strategies on AID\cite{2017AID} and NR45\cite{2017Remote}.}
\vspace{-0.6cm}
\end{figure}
\begin{figure*}
\vspace{0.2cm}
	\centering 
	\includegraphics[width=7in]{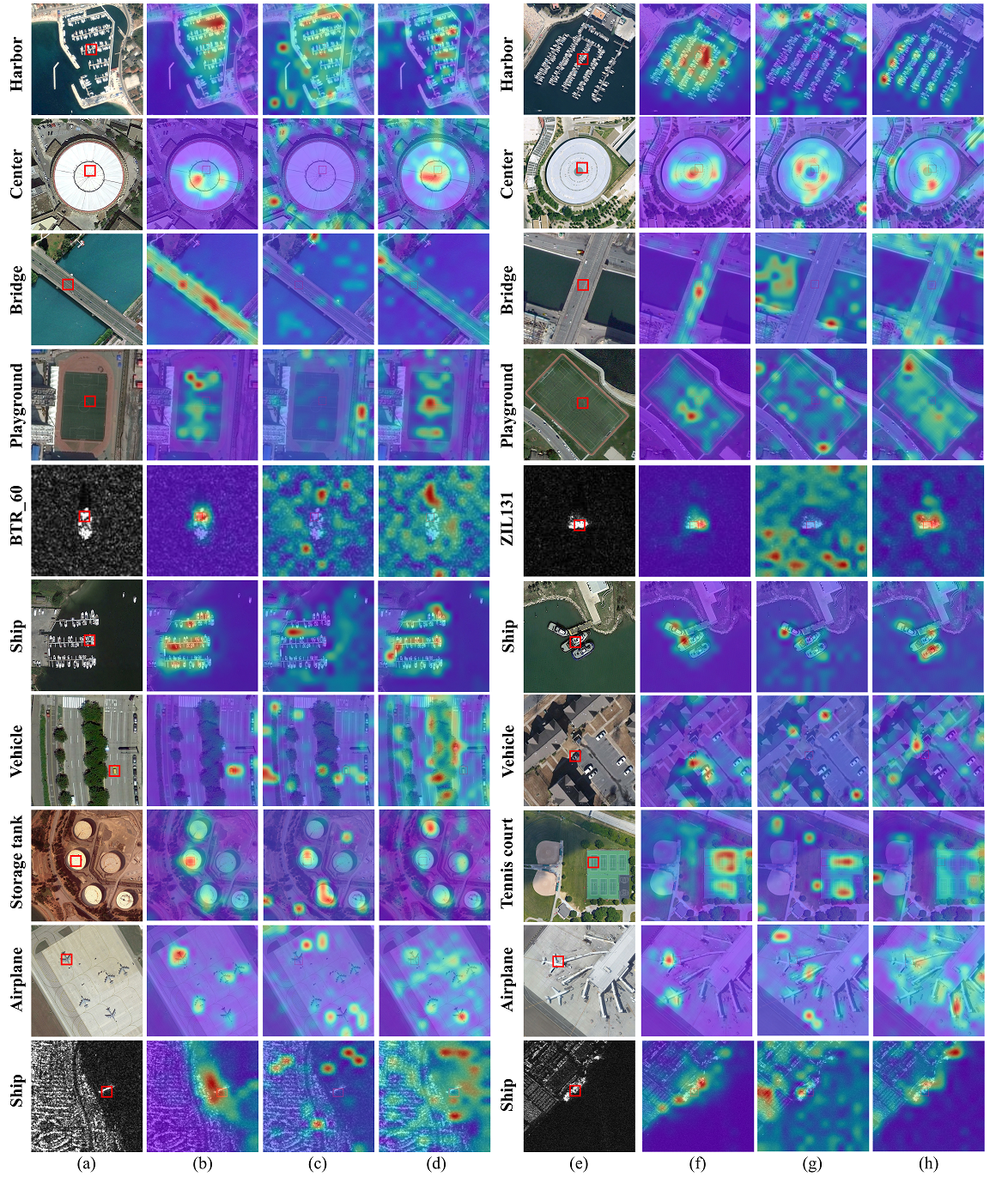}
	\caption{The comparison of self-attention maps. Column (a) and (e) represent original images which are from optical and SAR images; Column (b) and (f) are score maps produced by our proposed CSPT; Column (c) and (g) are score maps produced by supervised straightforward pretraining-then-fine-tuning; Column (d) and (h) denote score maps produced by self-supervised straightforward pretraining-then-fine-tuning.}
\vspace{-0.6cm}
\end{figure*}
    \par Subsequently, as discussed in Section \uppercase\expandafter{\romannumeral3} A, the existing domain gap between nature and remote sensing scenes would limit the performance of the fine-tuning step even for the model pretrained by SSP(IN1K). Consequently, the consecutive pretraining paradigm is designed to join more relevant unlabeled data (i.e., SSP(IN1K$\rightarrow$Train) and SSP(IN1K$\rightarrow$(Train+Test)) and wait for more iterative epochs (e.g., ep800, ep1600 and ep2400) in the further self-supervised pretraining step. This can bridge the domain gap between the pretrained model from IN1K\cite{ILSVRC15} and fine-tuning on remote sensing data (i.e., gradually bridge the domain gap between nature and remote sensing scenes). Then, as reported in the 8th, 9th and 10th columns of Table \uppercase\expandafter{\romannumeral2}, we can see that after self-supervised pretraining on IN1K (i.e., SSP(IN1K)), when the extra unlabeled training data are added into the further self-supervised pretraining step of SSP(IN1K$\rightarrow$Train) at the 800, 1600 and 2400 epochs, it can promote the performance of the fine-tuning step on all downstream tasks compared with other pretraining methods (i.e., SSP(IN1K), SP(M-AID) and SP(IN1K)). Next, according to the 4th and 8th columns of Table \uppercase\expandafter{\romannumeral2}, when ViT-B \cite{dosovitskiy2020image} is pretrained by the newly designed CSPT, it can obtain superior scene classification performance compared to SP(M-AID). This shows that consecutive self-supervised pretraining with extra relevant unlabeled data and more iterative epochs is a more promising method than building a large-scale remote sensing dataset for supervised pretraining. Moreover, to exhibit the superiority of narrowing the existing domain gap for the fine-tuning step via the designed CSPT, the Top1 accuracy curves of ViT-B\cite{dosovitskiy2020image} fine-tuning on AID\cite{2017AID} and NR45\cite{2017Remote} are plotted in Fig. 6 (a) and (b). Here, the green curves indicate the model convergence situation of SSP(IN1K$\rightarrow$Train) on AID\cite{2017AID} and NR45\cite{2017Remote}, and orange and blue curves individually represent the fine-tuning convergence trends of SP (IN1K) and SSP(IN1K) on AID\cite{2017AID} and NR45\cite{2017Remote}. These curves in Fig. 6 (a) and (b) indicate that the designed CSPT can facilitate the ViT-B\cite{dosovitskiy2020image} to rapidly converge and obtain the highest Top1 accuracy with fewer iterative epochs at the fine-tuning step. Following the core idea of not stopping pretraining, we consider enlarging the data volume on the further self-supervised pretraining step (i.e., utilize the relevant unlabeled data of training and testing sets) and wait for more epochs (e.g., ep800, ep1600, ep2400) for SSP(IN1K$\rightarrow$(Train+Test)). We can see that SSP(IN1K$\rightarrow$(Train+Test)) can further boost the fine-tuning step and  obtain the best performance on all downstream tasks, as reported in the 11th, 12th and 13th columns of Table \uppercase\expandafter{\romannumeral2}. Here, the best results are marked in bold, and we find that compared with pretraining method of SSP(IN1K), the designed CSPT can advance the fine-tuning step by 0.62\%$\sim$7.1\% improvements for all downstream tasks on the nine public optical remote sensing datasets. 
    \par Moreover, related to the problem analysis of domain gap in Section \uppercase\expandafter{\romannumeral3} A, several visualized attention score maps are shown in Fig. 7. We can see that the designed CSPT can easily adapt to any category of imaging data (e.g., optical RGB images or SAR) and correctly identify the selected area (i.e., the red areas with higher attention scores) from the original images of (a) and (e) in Fig. 7. For example, compared with the attention score maps of (c) and (g) produced by SP(IN1K) or (d) and (h) produced by SSP(IN1K), the attention score maps of SSP(IN1K$\rightarrow$(Train+Test)) have higher attention scores located in relevant areas and fewer attention scores distributed in irrelevant areas. In general, the designed CSPT is a promising task-aware model training paradigm in RSD than previous methods. It can convert the expensive manual labeling of large-scale datasets into an open selection of relevant unlabeled data applied to the further self-supervised pretraining step to bridge the domain gap. In addition, it only needs to wait for more iterative epochs to transfer the domain-level knowledge into RSD. Furthermore, based on the reported results in Table \uppercase\expandafter{\romannumeral2}, we can see that the MIM-based pretext task of task-agnostic representation applied in the designed CSPT is friendly with diverse downstream tasks because it mitigate the task-aware discrepancy and largely promote the performance of fine-tuning for diverse downstream tasks.
\begin{table*}[]
\vspace{0.2cm}
\renewcommand\arraystretch{1.2}
\caption{The comparison of fine-tuning on few labeled samples.}
\centering
\setlength{\tabcolsep}{12pt}{
\begin{tabular}{cccccccccc}
\toprule[1.2pt]
\multirow{2}{*}{Pretraining Method}      & \multirow{2}{*}{Architecture} & \multicolumn{4}{c}{NR45}      & \multicolumn{4}{c}{AID}      \\ \cline{3-10}
                              &                               & 2\%   & 4\%   & 8\%   & 16\%  & 2\%   & 4\%   & 8\%   & 16\%  \\ \toprule[1pt]
\multirow{2}{*}{Train from scratch} & ResNet-50                      & 27.48 & 44.30 & 54.62 & 63.73 & 21.80 & 26.11 & 46.15 & 59.78 \\
                              & ViT-B                     & 27.50 & 38.73 & 51.05 & 66.19 & 23.17 & 32.05 & 43.97 & 59.65 \\ \hline
\multirow{2}{*}{SP(IN1K)}   & ResNet-50                      & 73.45 & 80.92 & 86.77 & 91.34 & \textbf{65.60} & \textbf{75.70} & 86.10 & 92.62 \\
                              & ViT-B                     & 75.80 & 81.51 & 90.24 & 93.08 & 39.03 & 66.72 & 88.19 & 93.47 \\ \hline
SSP(IN1K)                  & ViT-B                     & 59.36 & 85.92 & 88.38 & 92.58 & 17.59 & 42.38 & 83.29 & 93.44 \\ \hline
CSPT                      & ViT-B                     & \textbf{80.43} & \textbf{89.66} & \textbf{92.56} & \textbf{94.33} & 42.79 & 73.69 & \textbf{93.23} & \textbf{96.10} \\ \toprule[1.2pt]
\end{tabular}
}
\vspace{-0.6cm}
\end{table*}
\begin{figure}
\vspace{0.2cm}
	\centering 
	\includegraphics[width=3.5in]{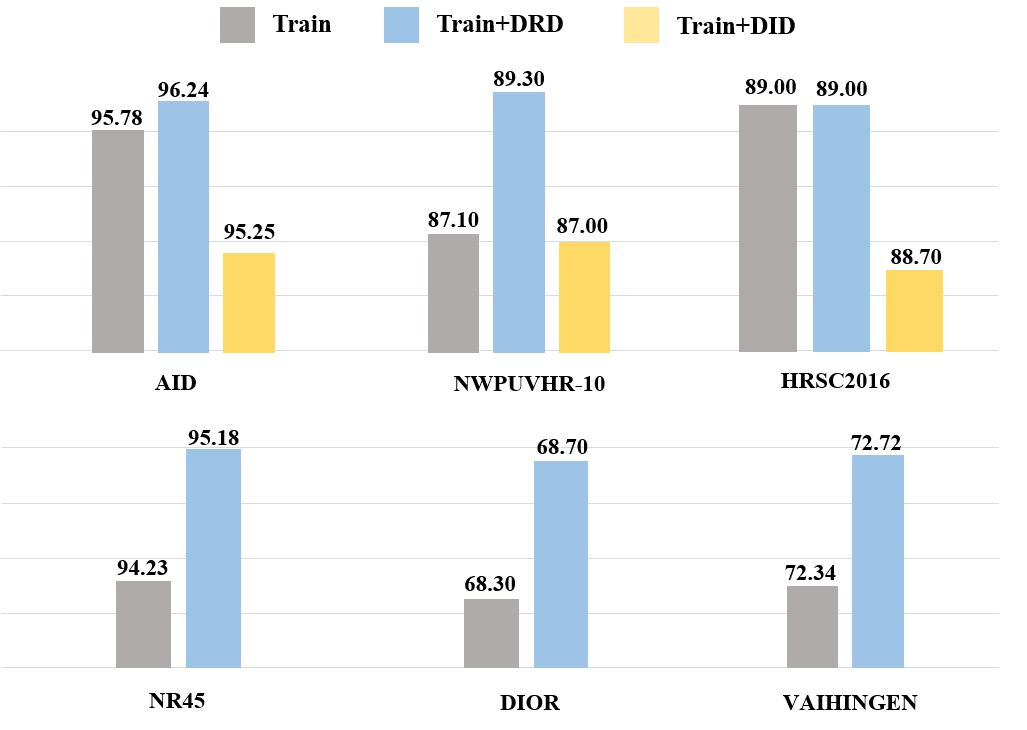}
	\caption{Discussion on different data expansion on the further self-supervised pretraining step. Notably, AID\cite{2017AID} and NR45\cite{2017Remote} are reported in mean average of Top-1 classification accuracy; NWPUVHR-10\cite{cheng2014multi}, HRSC2016\cite{liu2017high} and DIOR\cite{2019Object} are reported in mAP@0.5; VAIHINGEN\cite{isprs} is reported in mIoU.}
\vspace{-0.6cm}
\end{figure}
\subsection{Scalability of Data Volume}
    According to Section \uppercase\expandafter{\romannumeral4} C, the further self-supervised pretraining step of the designed CSPT is very important for the knowledge transfer learning. In addition, we also find that in Table \uppercase\expandafter{\romannumeral2}, the different unlabeled data volumes applied in the further self-supervised pretraining step can impact the fine-tuning performance. Thus, here, the scalability of data volume is discussed. Specifically, we formulate two unlabeled data expansion settings, namely,  (1) domain relevant data (DRD) and (2) domain irrelevant data (DID) to objectively analyze the impact of different unlabeled data joining in the further self-supervised pretraining step. In addition, from the view of fine-tuning step, RSD generally possesses insufficient data volume for fine-tuning on low-resource downstream tasks, as shown in Table \uppercase\expandafter{\romannumeral1}. Thus, to verify that our proposed CSPT can also adapt to low-resource downstream tasks, we further analyze the impact of fewer labeled samples applied in the fine-tuning step under different pretraining methods.

\subsubsection{Discussion on Further Self-supervised Pretraining Step}  
    As shown in Fig. 8, to verify the impact of adding more relevant unlabeled data for further self-supervised pretraining step, we set up a large-scale unlabeled dataset (i.e., DRD) by gathering three dataset images in RSD, including DOTA \cite{xia2018dota}, DIOR \cite{2019Object} and NR45 \cite{2017Remote}, which contains 66,593 images and covers almost all kinds of remote sensing images with various scenes. Next, as the blue and gray bars show in Fig. 8, through further self-supervised pretraining for 800 epochs on the combination of an unlabeled training set of diverse downstream tasks and DRD, these fine-tuning results obtain more gain than the gain obtained by only using unlabeled training data of given downstream tasks. The reason is that the DRD involves relevant unlabeled data of most downstream tasks. Thus, a comprehensive data distribution can facilitate the generalist model from IN1K\cite{ILSVRC15} to further learn the general knowledge representation for RSD, which can then easily adapt to diverse downstream tasks. 
    \par As for the data expansion setting (2), the impact of expanding DID is discussed. Referring to the training sets of downstream datasets (e.g., AID \cite{2017AID}, NWPUVHR-10\cite{cheng2014multi} and HRSC2016 \cite{liu2017high}), the same amount of unlabeled data from natural scene dataset of Place365 \cite{zhou2017places} is considered to be DID. Then, it is combined with given training sets for further self-supervised pretraining with 800 epochs. As shown in the yellow and gray bars reported in Fig. 8, we find that expanding the unlabeled DID would cause accuracy reductions of 0.53\%, 0.1\% and 0.3\% on AID\cite{2017AID}, NWPUVHR-10\cite{cheng2014multi} and HRSC2016\cite{liu2017high}, respectively. This further demonstrates the existence of a domain gap between nature and remote sensing scenes. In general, based on the above analysis, when simply taking data correlation into account, we conclude that consecutive self-supervised pretraining has the ability to release a huge potential of unlabeled data for promoting the fine-tuning performance in RSD.
\subsubsection{Discussion on Fine-tuning Step}
\begin{figure*}
\vspace{0.2cm}
	\centering 
	\includegraphics[width=6in]{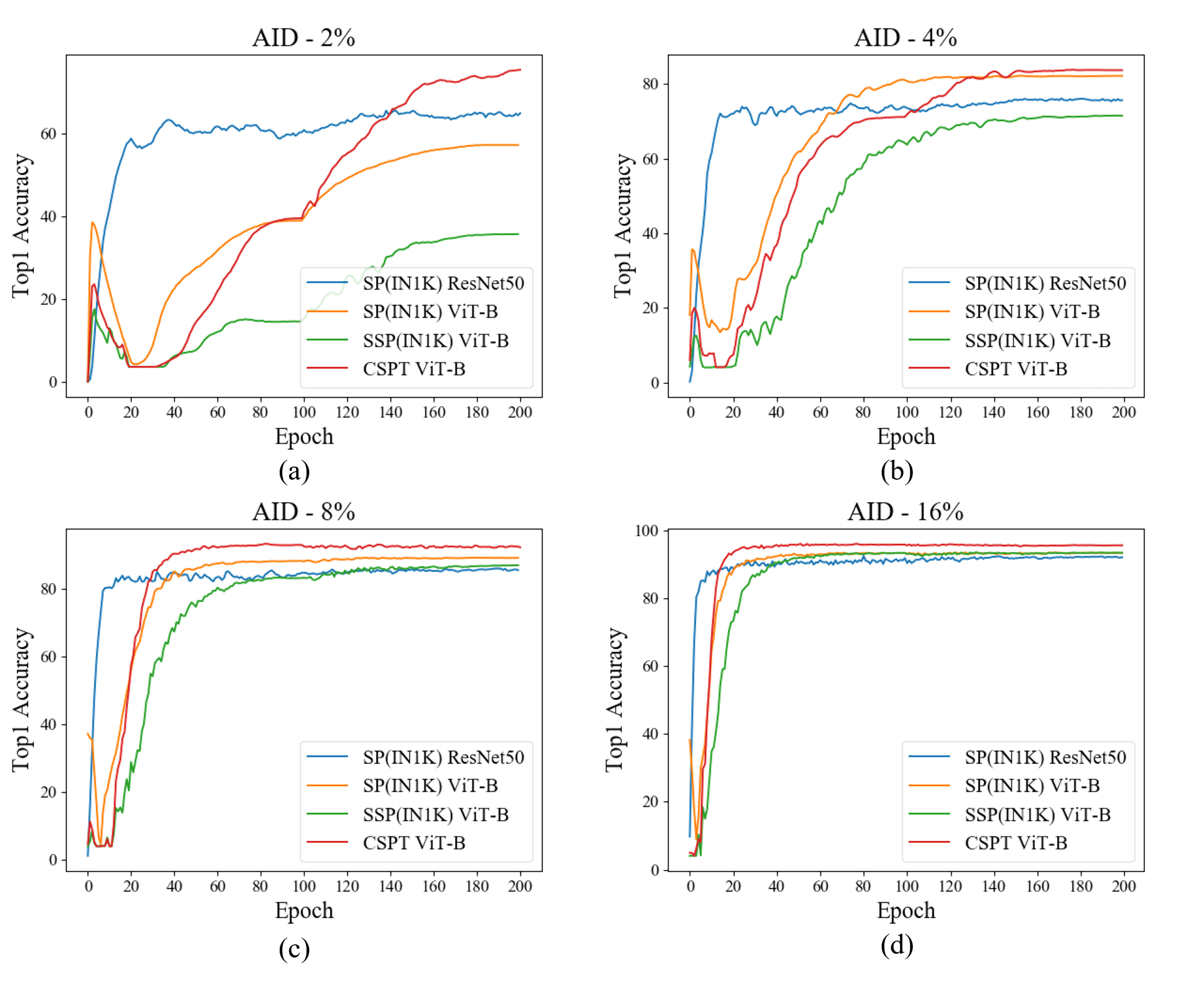}
	\caption{The Top.1 accuracy curves of fine-tuning on 2\%, 4\%, 8\%, 16\% of AID\cite{2017AID}.}
\vspace{-0.6cm}
\end{figure*}
    As shown in Table \uppercase\expandafter{\romannumeral3}, to explore the scalability of our proposed CSPT on low-resource downstream tasks, different pretraining methods (e.g., train from scratch, SP(IN1K) and SSP(IN1K)) with different network architectures (e.g., ViT-B\cite{dosovitskiy2020image} and ResNet-50\cite{he2016deep}) are selected for comparison. Notably, in our designed CSPT setting, all images of AID\cite{2017AID} and NR45\cite{2017Remote} can be regarded as task-relevant unlabeled data and adopted for the further self-supervised pretraining step. Then, for fair comparison , we fine-tuned all pretrained models on the training set ratios of 2\%, 4\%, 8\%, 16\% on AID\cite{2017AID} and NR45\cite{2017Remote} for 100 epochs (default setting). According to Table \uppercase\expandafter{\romannumeral3}, one observation is that our proposed CSPT produces the best results in all training set ratios of NR45 \cite{2017Remote} and 8\% and 16\% training set ratios of AID\cite{2017AID}. However it only obtains suboptimal results on 2\% and 4\% training set ratios of AID \cite{2017AID}. It is worthy noting that from Table \uppercase\expandafter{\romannumeral1}, AID\cite{2017AID} has less data volume than NR45\cite{2017Remote} so that when using 2\% or 4\% training set ratios of AID\cite{2017AID}, there are 5$\sim$9 samples per class used for training of the fine-tuning step. Thus, our proposed CSPT for ViT-B\cite{dosovitskiy2020image} does not seem to handle well fewer labeled samples. However, by analyzing the accuracy trends shown in Fig. 9, we observe that the blue curve of SP(IN1K) for ResNet-50\cite{he2016deep} reaches the performance bottleneck quickly, while the red curve of our proposed CSPT for ViT-B\cite{dosovitskiy2020image} has been growing in the default epoch setting (i.e., 100 epochs). Thus, the number of epochs is expanded to explore the performance bottleneck of our proposed CSPT for ViT-B\cite{dosovitskiy2020image}. As shown in Fig. 9, the blue curves remain at a constant performance level, while the red curves exceed the blue curves and reach the best performance with the increase in epochs. We conclude that differences in convergence speed are rooted in network architecture. Because ViT \cite{dosovitskiy2020image} has much less image-specific inductive biases than CNNs, it requires a longer time to learn the relevant patterns from fewer labeled samples. However, according to Fig. 9 (a), with the increase in the number of training epochs, the orange and green curves of SP(IN1K) and SSP(IN1K) for ViT-B\cite{dosovitskiy2020image} still do not exceed the blue curve of SP(IN1K) for ResNet-50\cite{he2016deep}. This demonstrates that our proposed CSPT can accelerate the convergence of ViT\cite{dosovitskiy2020image} and bring performance improvement on low-resource downstream tasks by learning ahead task-aware representation knowledge from unlabeled images.

\subsection{Scalability on SAR Imaging Data}
    We notice that image data in RSD possess the characteristics of multi-payload and multi-platform. Thus, apart from optical remote sensing images, SAR satellites are also essential payloads. However, the number of available SAR images is limited, and the SAR images have a larger domain gap with natural scene images. Thus, to verify the scalability of our proposed CSPT, we also discuss the performance of the designed CSPT strategy on SAR images. Following the same experimental setting, we conduct experiments on one target classification dataset, MSTAR \cite{1998Moving}, and two ship detection datasets, HRSID \cite{wei2020hrsid} and SSDD \cite{li2017ship}. Their descriptions are listed in Table \uppercase\expandafter{\romannumeral1}. In detail, the unlabeled training and training+testing data of given datasets are adopted for the further self-supervised pretraining step of the designed CSPT with 800,1600 and 2400 epochs, respectively. The experimental results are reported in Table \uppercase\expandafter{\romannumeral4}. Compared with the straightforward self-supervised pretrain-then-finetune method (i.e., see the 3rd column of Table \uppercase\expandafter{\romannumeral4}), the designed CSPT (i.e., IN1K$\rightarrow$(Train+Test)) boosts the accuracy by 1.94\% on MSTAR\cite{1998Moving}, 0.9\% mAP on SSDD\cite{li2017ship} and 1.5\% mAP on HRSID\cite{wei2020hrsid}. This demonstrates that our proposed CSPT can effectively transfer domain-level knowledge of large-scale nature scene data into downstream tasks for SAR images too.
\vspace{-0.1cm}

\begin{table*}[]
\vspace{0.2cm}
\renewcommand\arraystretch{1.2}
\caption{The performance of CSPT strategy on SAR images.}
\centering
\begin{threeparttable}
\setlength{\tabcolsep}{13pt}{
\begin{tabular}{ccccccccc}
\toprule[1.1pt]
\multirow{3}{*}{Task}             & \multirow{3}{*}{Dataset} & \multicolumn{7}{c}{Pretraining Setting} \\ \cline{3-9}
                                   &   &                         IN1K  & \multicolumn{3}{c}{IN1K$\rightarrow$Train} & \multicolumn{3}{c}{IN1K$\rightarrow$(Train+Test)} \\ \cmidrule(r){3-3} \cmidrule(r){4-6} \cmidrule(r){7-9} 
                                  &                          & ep800 & ep800     & ep1600    & ep2400   & ep800     & ep1600    & ep2400   \\ \toprule[1pt]
Target   Classification           & MSTAR\cite{1998Moving}                    & 98.03 & 99.82     & 99.80     & 99.80    & 99.96     & 99.96 & \textbf{99.97}     \\ \hline
\multirow{2}{*}{Ship   Detection} & SSDD\cite{li2017ship}                     & 90.90 & 91.80     & 91.50     & 91.30    & 91.00     & 91.70     & \textbf{91.80}   \\
                                  & HRSID \cite{wei2020hrsid}                   & 68.70 & 68.90     & 69.10     & 69.90    & 69.00     & 69.70     & \textbf{70.20}   \\ \toprule[1.1pt]
\end{tabular}}
\begin{tablenotes}
    \footnotesize
    \item *Note: 1) Network architecture uses ViT-B \cite{dosovitskiy2020image}. 2) Evaluation metric: the average of Top-1 classification accuracy for target classification; the average precision (AP@0.5) for ship detection.
\end{tablenotes}
\end{threeparttable}
\vspace{-0.6cm}
\end{table*}
\begin{table*}[]
\vspace{0.1cm}
\caption{The comparison results on AID\cite{2017AID} and NR45\cite{2017Remote}. }
\renewcommand\arraystretch{1.2}
\centering
\setlength{\tabcolsep}{13pt}{\begin{tabular}{cccccc}
\toprule[1.1pt]
Method              &Year          & Setting                                            & Network               & AID(2:8) & NR45(2:8) \\ \toprule[1pt]
MG-CAP\cite{MG-CAPTIP2020}              &TIP2020       & SP(IN1K)                                             & VGG-16                 & 93.34    & 92.95     \\
CAD\cite{CADJSTAR2020}                 &JSTAR2020     & SP(IN1K)                                             & DenseNet-121           & 95.73    & 94.58     \\
KFBNet\cite{KFBNetTGRS2020}              &TGRS2020      & SP(IN1K)                                             & DenseNet-121           & 95.50    & 95.11     \\
F2BRBM\cite{F2BRBMJSTAR2021}              &JSTAR2021     & SP(IN1K)                                             & ResNet-50              & 96.05    & 94.87     \\
MBLANet\cite{MBLANetTIP2021}             &TIP2021       & SP(IN1K)                                             & ResNet-50              & 95.60    & 94.66     \\
EAM\cite{EAMGRSL2020}                 &GRSL2021      & SP(IN1K)                                             & ResNet-101             & 94.26    & 94.29     \\
MSA-Net\cite{MSANetJSTAR2021}             &JSTAR2021     & SP(IN1K)                                             & ResNet-101             & 93.53    & 93.52     \\
ESD-MBENet\cite{ESD-MBENet}          &TGRS2021      & SP(IN1K)                                             & DenseNet-121           & 96.39    & 95.36     \\
ASP\cite{long2022aerial}                 &arXiv2022     & SP(M-AID)                                          & ResNet-101             & 95.40    & 94.20     \\
SeCo\cite{SeCo}                &ICCV2021      & SSP(Sentinel-2)                                            & ResNet-50              & 93.47    & 92.91     \\
MoCo\cite{MoCo}                &ICCV2021      & SSP(IN1K)                                            & ResNet-50              & 92.51    & 91.79     \\ \hline
Swin Transformer\cite{liu2021swin}    &ICCV2021      & SP(IN1K)                                             & Swin-T                & 96.55    & 94.70     \\
Vision Transformer\cite{dosovitskiy2020image}  &ICLR2021      & SP(IN1K)                                             & ViT-B             & 94.04    & 94.10     \\
CTNet\cite{CTNet}               &GRSL2021      & SP(IN1K)                                             & MobileNet-v2+ViT-B     & 96.25    & 95.40     \\
ViTAEv2\cite{ViTAEv2}             &arXiv2022     & SP(IN1K)                                             & ViTAEv2-S             & 96.61    & 95.29    \\
RSP\cite{wang2022empirical}                 &arXiv2022     & SP(M-AID)                                          & ViTAEv2-S-E40         & 96.72    & 95.35     \\
RSP\cite{wang2022empirical}                 &arXiv2022     & SP(M-AID)                                          & ViTAEv2-S-E100        & \textbf{96.91}    & 95.60     \\
SimMIM\cite{xie2021simmim}              &CVPR2022      & SSP(IN1K)                                            & ViT-B             & 93.08    & 92.57     \\
MAE\cite{he2021masked}                 &CVPR2022      & SSP(IN1K)                                            & ViT-B             & 95.00    & 93.94     \\
MAE\cite{he2021masked}                 &CVPR2022      & SSP(AID or NR45)                                            & ViT-B             & 86.53    & 88.40     \\
MAE\cite{he2021masked}                 &CVPR2022      & SSP(IN1K)                                            & ViT-L             & 94.92    & 94.34     \\
CSPT                 &Our             & SSP(IN1K$\rightarrow$Train+DRD)                            & ViT-B             & 96.24    & 95.18     \\
CSPT                 &Our             & SSP(IN1K$\rightarrow$Train+Test)                     & ViT-B             & 96.75    & 95.11     \\
CSPT                 &Our             & SSP(IN1K$\rightarrow$Train+Test)                     & ViT-L             & 96.30    &\textbf{95.62}  \\   \toprule[1.1pt]
\end{tabular}}
\vspace{-0.6cm}
\end{table*}
\begin{table*}[]
\vspace{0.2cm}
\Large
\renewcommand\arraystretch{1.2}
\caption{The comparison results on DIOR\cite{2019Object}.}
\centering
\resizebox{\textwidth}{30mm}{\begin{tabular}{cccccccccccccccccccccccc}
\toprule[2pt]
Method      & Setting           & Backbone                                  & mAP  & AL   & AT   & BF   & BC   & B    & C    & D    & ESA  & ETS  & GC   & GTF  & HB   & O    & S    & SD   & ST   & TC   & TS   & V    & WM   \\ \toprule[1pt]
Faster-RCNN\cite{ren2015faster} & SP(IN1K)    & ResNet-101                                       & 53.6 & 51.3 & 61.6 & 62.2 & 80.6 & 26.9 & 74.2 & 37.3 & 53.4 & 45.1 & 69.6 & 61.8 & 43.7 & 48.9 & 56.1 & 41.8 & 39.5 & 73.8 & 44.7 & 33.9 & 65.3 \\
Mask-RCNN\cite{maskrcnn_he2017}   & SP(IN1K)     & ResNet-101                                       & 65.2 & 53.9 & 76.6 & 63.2 & 80.9 & 40.2 & 72.5 & 60.4 & 76.3 & 62.5 & 76.0 & 75.9 & 46.5 & 57.4 & 71.8 & 68.3 & 53.7 & 81.0 & 62.3 & 53.0 & 81.0 \\
YOLOv5\cite{yolov5}      & SP(IN1K)     & CSPdarknet-53                                    & 68.5 & 87.3 & 61.7 & 73.7 & 90.0 & 42.6 & 77.5 & 55.2 & 63.8 & 63.2 & 66.9 & 78.0 & 58.1 & 58.1 & 87.8 & 54.3 & \textbf{79.3} & 89.7 & 50.2 & 53.9 & 79.6 \\
CenterNet\cite{CenterNet_zhou2019} & SP(IN1K)     & DLA-34                                           & 63.2 & 78.6 & 56.5 & 76.1 & 88.1 & 33.2 & 77.1 & 41.0 & 47.4 & 55.5 & 71.4 & 72.5 & 23.0 & 52.7 & \textbf{89.8} & 54.0 & 78.6 & 86.2 & 46.1 & \textbf{57.8} & 77.4 \\
CANet\cite{CANet_JSTAR2020}       & SP(IN1K)    & ResNet-101                                       & \textbf{74.3} & 70.3 & 82.4 & 72.0 & 87.8 & \textbf{55.7} & 79.9 & 67.7 & 83.5 & \textbf{77.2} & 77.3 & \textbf{83.6} & 56.0 & \textbf{63.6} & 81.0 & 79.8 & 70.8 & 88.2 & 67.6 & 51.2 & 89.6 \\
PANet\cite{PANet} & SP(IN1K)    & ResNet-101                                       & 66.1 & 60.2 & 72.0 & 70.6 & 80.5 & 43.6 & 72.3 & 61.4 & 72.1 & 66.7 & 72.0 & 73.4 & 45.3 & 56.9 & 71.7 & 70.4 & 62.0 & 80.9 & 57.0 & 47.2 & 84.5 \\
MSFC-Net\cite{msfcnet}    & SP(IN1K)    & ResNeSt-101                                      & 70.1 & 85.8 & 76.2 & 74.3 & 90.1 & 44.1 & 78.1 & 55.5 & 60.9 & 59.5 & 76.9 & 73.6 & 49.5 & 57.2 & 89.6 & 69.2 & 76.5 & 86.7 & 51.8 & 55.2 & 84.3 \\
FSoD\cite{wang2021fsod}        & SP(NR45)  & MSE-Net                                         & 71.8 & \textbf{88.9} & 66.9 & \textbf{86.8} & \textbf{90.2} & 45.5 & 79.6 & 48.2 & 86.9 & 75.5 & 67.0 & 77.3 & 53.6 & 59.7 & 78.3 & 69.9 & 75.0 & \textbf{91.4} & 52.3 & 52.0 & \textbf{90.6} \\ \hline
Mask-RCNN(MAE)\cite{he2021masked}  & SSP(IN1K)    & ViT-B                                       & 66.8 & 58.9 & 85.6 & 69.4 & 80.7 & 37.8 & 78.5 & 70.2 & 85.0 & 55.4 & 80.7 & 77.4 & 58.7 & 57.1 & 44.3 & 79.2 & 44.3 & 83.1 & 70.9 & 27.5 & 74.8 \\
Mask-RCNN(MAE)\cite{he2021masked}  & SSP(IN1K)    & ViT-L                                       & 68.3 & 66.1 & 86.5 & 73.3 & 83.6 & 41.4 & 81.6 & 72.2 & 86.2 & 58.3 & 79.2 & 78.7 & 60.3 & 61.1 & 60.1 & 73.4 & 42.1 & 83.3 & 71.3 & 28.9 & 78.7 \\
Mask-RCNN(MoCo)\cite{MoCo}  & SSP(IN1K)   & ResNet-50                                    & 62.5 & 57.9 & 75.1 & 65.1 & 85.3 & 36.2 & 71.9 & 59.2 & 66.4 & 51.6 & 74.0 & 75.8 & 58.8 & 54.8 & 67.8 & 67.8 & 44.2 & 83.0 & 58.4 & 27.6 & 76.6 \\
Mask-RCNN(SimMIM)\cite{xie2021simmim} & SSP(IN1K)   & ResNet-50                                   & 63.5 & 59.6 & 80.4 & 69.7 & 77.0 & 34.5 & 77.5 & 64.9 & 77.6 & 52.4 & 76.8 & 74.4 & 52.0 & 55.5 & 59.6 & 70.8 & 40.5 & 80.2 & 64.4 & 27.1 & 75.0 \\
Mask-RCNN(Our)  & SSP(IN1K$\rightarrow$Train+DRD)  & ViT-B                        & 68.7 & 69.9 & 87.7 & 70.8 & 81.2 & 41.6 & 80.5 & 74.8 & 86.0 & 58.8 & 78.9 & 75.6 & 60.6 & 58.9 & 60.8 & 78.3 & 44.6 & 84.1 & 76.2 & 29.0 & 76.4 \\
Mask-RCNN(Our) & SSP(IN1K$\rightarrow$Train+Test) & ViT-B                   & 69.8 & 69.8 & 89.1 & 74.7 & 82.6 & 42.2 & 80.5 & \textbf{76.9} & 86.4 & 58.8 & 80.7 & 77.7 & \textbf{61.9} & 60.2 & 60.9 & 79.2 & 46.1 & 84.3 & 77.2 & 29.0 & 77.3 \\
Mask-RCNN(Our) & SSP(IN1K$\rightarrow$Train+Test) & ViT-L                   & 71.7 & 74.1 & \textbf{89.9} & 81.2 & 86.2 & 44.5 & \textbf{81.9} & 74.8 & \textbf{90.1} & 61.3 & \textbf{81.9} & 79.6 & 61.6 & 61.0 & 61.0 & \textbf{83.7} & 44.5 & 88.1 & \textbf{78.9} & 29.2 & 79.9 \\ \toprule[2pt]
\end{tabular}}
\vspace{-0.6cm}
\end{table*}

\begin{table}[]
\vspace{0.2cm}
\caption{The comparison results on ISPRS Potsdam\cite{isprs}. }
\renewcommand\arraystretch{1.2}
\centering
\setlength{\tabcolsep}{4pt}{
\begin{tabular}{cccc}
\toprule[1.2pt]
Method       & Setting              & Backbone  & mIoU  \\ \toprule[1pt]
BES-Net\cite{chen2022bes}      & SP(IN1K)             & ResNet-18 & 78.21 \\
Deeplabv3+\cite{deeplabv3+}   & SP(IN1K)             & ResNet-50  & 75.21 \\
Upernet\cite{xiao2018upernet}      & SP(IN1K)             & ResNet-50  &   75.86    \\
GCNet\cite{cao2019gcnet}        & SP(IN1K)             & ResNet-101 &  75.38     \\
Upernet\cite{xiao2018upernet}      & SP(IN1K)             & ViT-B     & 76.43 \\ \hline
Upernet(Our) & SSP(IN1K$\rightarrow$Train)      & ViT-B     & 78.36 \\
Upernet(Our) & SSP(IN1K$\rightarrow$Train+Test) & ViT-B     & \textbf{78.70} \\ \toprule[1.2pt]
\end{tabular}}
\vspace{-0.6cm}
\end{table}
\subsection{Comparison Experiment Analysis}
    In this section, to demonstrate the potential and superiority of the designed CSPT, we apply our proposed CSPT to train plain networks in RSD to compare with advanced methods. Therefore, recently proposed SOTA methods and outstanding pretraining technologies are selected and compared on four public remote sensing datasets (i.e., AID\cite{2017AID}, NR45\cite{2017Remote}, DIOR\cite{2019Object} and ISPRS Potsdam\cite{isprs}) involving three downstream tasks (e.g., scene classification, object detection and land cover classification).
\subsubsection{Scene Classification}
    Following \cite{long2022aerial}, the unified data division is adopted to evaluate AID\cite{2017AID} and NR45\cite{2017Remote}. In Table \uppercase\expandafter{\romannumeral5}, the best results are marked in bold, and different pretraining methods and networks of these comparison methods are listed in the 3rd and 4th columns. Firstly, many well-designed modules have been adopted for improving performance to achieve SOTA results, such as advanced attention mechanisms (e.g., CAD \cite{CADJSTAR2020}, EAM \cite{EAMGRSL2020}, MBLANet \cite{MBLANetTIP2021}, ESD-MBENet \cite{ESD-MBENet} and MSA-Net\cite{MSANetJSTAR2021}) and powerful feature fusion modules (e.g., MG-CAP \cite{MG-CAPTIP2020}, F2BRBM\cite{F2BRBMJSTAR2021} and KFB-Net\cite{KFBNetTGRS2020}). From the experimental results of Table \uppercase\expandafter{\romannumeral5}, these carefully designed models have difficulty obtaining competitive results. Second, except for special module designs, some works also focus on studying powerful pretraining technologies. For example, MoCo\cite{MoCo}, MAE\cite{he2021masked} and SimMIM\cite{xie2021simmim} are designed for self-supervised pretraining on natural scene data. Here, we tried to transfer these pretrained models from large-scale nature scene data into the AID\cite{2017AID} and NR45\cite{2017Remote} datasets of RSD. From the 11th, 18th and 19th rows of Table \uppercase\expandafter{\romannumeral5}, it can be seen that MoCo\cite{MoCo} based on contrastive learning (CL) obtains slightly worse results than MAE\cite{he2021masked} and SimMIM\cite{xie2021simmim}. This is so since CL is decision-making based on deep semantic features that would lose much information from original images. In addition, CL needs to carefully construct positive and negative sample pairs; Otherwise, unsuitable positive and negative sample pairs would affect the performance of CL \cite{guo2022hcsc,peng2022crafting,li2022global}. In contrast, MIM-based pretext task applied for MAE \cite{he2021masked} and SimMIM\cite{xie2021simmim} directly reconstructs the pixels of the original images, which can not only perceive general and detailed image information but also avoid positive and negative sample allocation. Then, the reconstruction of randomly masked tokens is beneficial for evoking the cognition of underlying knowledge from images. In addition, to avoid the domain gap, some researchers have used self-built large-scale remote sensing data to pretrain their models (e.g., SeCo \cite{SeCo}, ASP\cite{long2022aerial} and RSP\cite{wang2022empirical}). From the 9th to 11th, 17th, 18th and 19th rows of Table \uppercase\expandafter{\romannumeral5}, it can be seen that these methods pretrained on the RSD dataset mostly suppress the above methods pretrained on nature scene dataset, which illustrates that domain gap would restrict the performance supremum. Although these methods achieve good performance, the expensive labor cost of the self-built large-scale dataset is inevitable, especially for datasets with manual labels. Meanwhile, it will also have the problem of label noise. Here, under the advanced self-supervised pretraining technology based on MIM task, we also tried to only pretrain on the AID\cite{2017AID} or NR45\cite{2017Remote} dataset to avoid the expensive collection cost but found that it achieved the worse performance of 86.53\% and 88.40\% as shown in the 20th row of Table \uppercase\expandafter{\romannumeral5}. This shows that the universal and transferable representation learned from large-scale data for the fine-tuning step is essential. Finally, according to the results of our proposed CSPT as shown in the 23rd and 24th rows, it obtained a promising result of 96.75\% on AID\cite{2017AID} and an SOTA result of 95.62\% on NR45\cite{2017Remote}. This indicates that the effectiveness of transferring nature scene domain-level knowledge into RSD by consecutive self-supervised pretraining.
\subsubsection{Object Detection}
    Object detection belongs to the dense prediction task, which has more complex network structure, including backbone, neck and head networks. Here, we use the Mask-RCNN \cite{maskrcnn_he2017} as the baseline model and then replace the backbone network with ViT-B/L \cite{dosovitskiy2020image} pretrained by our proposed CSPT. In addition, the public remote sensing dataset DIOR \cite{2019Object} is adopted to evaluate the performance. It contains 23,463 images with 192,472 instances involving 20 object categories, such as airplane (AL), airport (AT), baseball field (BF), basketball court (BC), bridge (B), chimney (C), dam (D), expressway service area (ESA), expressway toll station (ETS), golf course (GC), ground track field (GTF) harbor (HB), overpass (O), ship (S), stadium (SD), storage tank (ST), tennis court (TC), train station (TS), vehicle (V), and windmill (W). According to the data division rule of DIOR \cite{2019Object}, we compared our strategy with other SOTA methods and advanced pretraining technologies. Among these methods, some classical detectors, such as Faster-RCNN \cite{ren2015faster}, YOLOv5 \cite{yolov5}, Mask-RCNN\cite{maskrcnn_he2017}, PANet\cite{PANet} and CenterNet\cite{CenterNet_zhou2019} are selected. In addition, we also compare with some SOTA detectors from RSD including MSFC-Net \cite{msfcnet}, CANet\cite{CANet_JSTAR2020} and FSoD\cite{wang2021fsod}. 
    As reported in the 9th and 14th rows of Table \uppercase\expandafter{\romannumeral6}, compared with the baseline model Mask-RCNN \cite{maskrcnn_he2017}, our method achieves significant improvement (3\% mAP) without bells and whistles based on ViT-B \cite{dosovitskiy2020image}. In addition, when using ViT-L\cite{dosovitskiy2020image} without carefully designed modules, we obtain 71.7\% mAP, which is competitive with the SOTA results and even exceeds some of them.
    Moreover, some self-supervised pretraining methods (e.g., MAE\cite{he2021masked}, MoCo\cite{MoCo} and SimMIM\cite{xie2021simmim}) are compared as shown from the 9th to 12th rows of Table \uppercase\expandafter{\romannumeral6}, only using these pretrained models as backbones to implant into Mask-RCNN \cite{maskrcnn_he2017}. It can be observed that CSPT can facilitate the Mask-RCNN to achieve the best result.
\subsubsection{Land Cover Classification}
    For the land cover classification task, we adopt upernet\cite{xiao2018upernet} as the baseline model and then replace its backbone with the model pretrained by our proposed CSPT. Then, the ISPRS Potsdam \cite{isprs} is selected as our benchmark dataset to evaluate our proposed CSPT. As shown in Table \uppercase\expandafter{\romannumeral1}, there are six categories of land cover (e.g., impervious surface, building, low vegetation, tree, car and clutter) used for evaluating the performance (i.e., mIoU) of the models. Next, some advanced methods are selected for comparison such as Deeplabv3+\cite{deeplabv3+}, GCNet\cite{cao2019gcnet} and BES-Net\cite{chen2022bes} from nature and remote sensing scenes. As reported in the 5th and 7th rows of Table \uppercase\expandafter{\romannumeral7}, based on ViT-B \cite{dosovitskiy2020image}, our proposed CSPT brings 2.27\% mIoU gain compared with the baseline model. Moreover, our result is very competitive with other SOTA methods.
\section{Conclusions}
    In this paper, first, we provided an empirical analysis of knowledge transfer learning while illustrating some challenges (e.g., label noise and domain gap) of traditional knowledge transfer learning. Second, a concise and effective knowledge transfer learning strategy called CSPT based on the idea of not stopping pretraining in NLP is proposed to gradually narrow the domain gap and better transfer domain-level knowledge from the natural scene domain to the RSD. The further self-supervised pretraining step adopted in CSPT can release the potential of unlabeled data to facilitate fine-tuning on diverse downstream tasks in RSD. In addition, the MIM-based pretext task of task-agnostic representation that is utilized for self-supervised pretraining can mitigate the task-aware discrepancy from diverse downstream tasks. Finally, through extensive experiments, our proposed CSPT has been shown to bring significant performance improvements on various downstream tasks in RSD. Meanwhile, the comparison also shows that the designed CSPT can achieve the competitive results compared with SOTA methods on diverse downstream tasks without bells and whistles. In future work, we plan to explore more reasonable knowledge transfer strategies for specific downstream tasks.

\bibliographystyle{IEEEtran}
\bibliography{ref}

\end{document}